\documentclass[11pt]{article}
\usepackage[sort&compress,square,comma,authoryear]{natbib}
\usepackage[a4paper, total={6.1in, 10in}]{geometry}
\usepackage{enumitem}
\usepackage[utf8]{inputenc} 
\usepackage[T1]{fontenc}    
\usepackage{hyperref}       
\usepackage{url}            
\usepackage{booktabs}       
\usepackage{dirtytalk}
\usepackage{tabularx}
\usepackage{multirow}
\newcolumntype{Y}{>{\centering\arraybackslash}X}
\newcolumntype{Z}{>{\raggedleft\arraybackslash}X}
\usepackage[parfill]{parskip}
\usepackage{amsfonts}       
\usepackage{nicefrac}       
\usepackage{microtype}      
\usepackage[dvipsnames]{xcolor}         
\usepackage{amsmath,amssymb,amsthm}
\usepackage{algorithm,algorithmic}
\usepackage{bbold}
\usepackage{colortbl}
\usepackage{graphicx}
\usepackage{mathrsfs,mathtools}
\usepackage{subfigure}
\usepackage{thm-restate}
\usepackage{wrapfig}
\usepackage{caption}
\usepackage[capitalize]{cleveref}
\usepackage{tikz}
\usetikzlibrary{calc, fadings, decorations, shapes, positioning, arrows}
\usepackage[graphicx]{realboxes}
\usepackage{pifont}
\usepackage{tcolorbox}
\usepackage{hyperref}

\definecolor{salmon}{RGB}{234,153,153}
\definecolor{cornflowerblue}{RGB}{100,149,237}
\hypersetup{
    colorlinks,
    linkcolor={cornflowerblue},
    citecolor={cornflowerblue},
    urlcolor={salmon}
}

\definecolor{darkgreen}{rgb}{0.0, 0.5, 0.0}
\usepackage[textsize=tiny]{todonotes}
\usepackage{tabularx}
\usepackage{wrapfig}
\definecolor{darkblue}{rgb}{0, 0, 0.5}
\hypersetup{colorlinks=true, citecolor=darkblue, linkcolor=darkblue, urlcolor=darkblue}
\usepackage{tikz}

\theoremstyle{plain}
\newtheorem{theorem}{Theorem}[section]

\newtheorem{lemma}[theorem]{Lemma}

\theoremstyle{definition}
\newtheorem{definition}[theorem]{Definition}

\theoremstyle{remark}

\newcommand{\FGen}{T}
\newcommand{\pa}[1]{\mathsf{Pa}(#1)}
\newcommand{\expect}[1]{\mathbb{E}\left[#1\right]}
\newcommand{\pr}[2]{\mathbb{P}_{#1} ( #2 )}

\newcommand{\midsepremove}{\aboverulesep = 0mm \belowrulesep = 0mm}
\midsepremove


\title{AI Alignment in Medical Imaging: Unveiling Hidden Biases Through Counterfactual Analysis}

\date{}

\usepackage{authblk}

\author[1]{Haroui Ma$^{\dagger}$}
\author[4]{Francesco Quinzan\thanks{Part of this work was done while Francesco Quinzan was employed at the Department of Computer Science,  University of Oxford.}$^{\dagger}$}
\author[2,3]{Theresa Willem}
\author[1,3]{Stefan Bauer}
\affil[1]{TUM School of Computation, Information and Technology (CIT), Technical University Munich, Munich, Germany}
\affil[2]{TUM School for Medicine and Health, Institute of History and Ethics in Medicine, Technical University of Munich, Munich, Germany}
\affil[3]{Helmholtz AI, Helmholtz Munich, Munich, Germany}
\affil[4]{Department of Engineering Science, The University of Oxford, Oxford, UK}
\affil[ ]{\small \textit{${}^{\dagger}$These authors contributed equally to this work.}}
\affil[ ]{\small Corresponding author: francesco.quinzan@eng.ox.ac.uk}

\begin{document}

\maketitle

\begin{abstract}
\noindent
Machine learning (ML) systems for medical imaging have demonstrated remarkable diagnostic capabilities. However, their susceptibility to learning spurious correlations with sensitive attributes poses significant risks to fairness and safety. In this paper, we introduce a novel statistical framework to evaluate the dependency of medical imaging ML models on sensitive attributes, such as demographics. Our method leverages the concept of counterfactual invariance, measuring the extent to which a model's predictions remain unchanged under hypothetical changes to sensitive attributes. We present a practical algorithm that combines conditional latent diffusion models with statistical hypothesis testing to identify and quantify such biases without requiring direct access to counterfactual data. 
On synthetic benchmarks, our framework correctly identifies 96.1\% of biased models while producing false alarms on 22.0\% of fair models. On two real-world chest X-ray datasets, CheXpert and MIMIC-CXR, it detects bias at average rates 96.3\%, 95.7\% across all diagnostic tasks when models are strongly biased, and its average false alarm rate drops to 15.3\%, 14.7\%, respectively, for the least biased models, consistently outperforming existing fairness baselines and demonstrating strong alignment with counterfactual fairness principles.

\end{abstract}
\section{Author summary}
Machine learning (ML) systems are increasingly used in medical imaging for diagnostic purposes. However, these systems can learn biases from training data, leading to disparities in clinical outcomes, particularly across demographic groups. Our work addresses this critical issue by introducing a novel statistical framework that tests whether an ML model's predictions change when sensitive attributes---such as self-reported race or gender---are altered. We achieve this by using latent conditional diffusion models to create realistic image variations. These tools allow us to measure a model's dependency on sensitive attributes and assess its fairness. We validate our method on synthetic and real-world medical imaging datasets, demonstrating its effectiveness and alignment with counterfactual fairness principles. This approach represents an important step towards building trustworthy and unbiased AI systems in healthcare.

\section{Introduction}
\label{sec:introduction}
A major problem in using machine learning (ML) systems for healthcare is that these systems exhibit biases leading to disparities in performance for different groups of patients \citep{derma,derma2,lancet2,nature_med1}. For instance, patients from certain demographic groups may face a higher likelihood of misdiagnosis, such as false positives or false negatives. One possible reason for this problem is that AI models inadvertently learn and amplify existing confounding effects from the training data \citep{Banerjee2021AIRO}. 

With recent evidence that ML models can identify demographic factors such as age, sex, and racial identity from medical images \citep{Banerjee2021AIRO}---a skill beyond human capabilities---tracking confounding effects contained in medical imaging ML models is non-trivial \citep{rad1,rad2,rad3,rad4,rad5,Banerjee2021AIRO}. Although the exact mechanisms of demographic group membership prediction remain unclear \citep{Banerjee2021AIRO}, the observable capability suggests that AI models can learn to correctly correlate certain features from medical image data to real-world demographic attributes \citep{zengin2016ethnic}, relying on proxies or imaging-related surrogate covariates that correlate with specific attributes. These learned correlations, which may represent unfair treatment and discrimination in the real world, can distort a model's performance and induce harmful biases \citep{obermeyer2019dissecting}. When applied in clinical practice, models entailing such undetected biases may result in underdiagnosis and, thus, undertreatment \citep{rajpurkar2022ai}.

In this paper, we present a new framework to assess the dependency of imaging AI models' predictions on sensitive attributes, such as demographics. This framework allows us to evaluate whether a diagnostic model has learned to interpret certain visual features as indicators for sensitive attributes, and use these interpretations for its predictions, \emph{even if there is no causal relationship between these sensitive attributes and the development of a disease}. Our approach is grounded in the notion of counterfactual invariance \citep{DBLP:conf/nips/VeitchDYE21,DBLP:journals/corr/abs-2207-09768,DBLP:journals/corr/abs-2307-08519,DBLP:conf/clear2/FawkesES22}. Intuitively, counterfactual invariance measures the extent to which a model's prediction remains unchanged when an attribute is altered. Counterfactual invariance explicitly removes the natural dependence between $A$ and its causes, \emph{ensuring that the relationship between $A$ and the model's predictions is not confounded by factors upstream of $A$}.  

Our contribution lies in the design of a novel procedure that operationalizes counterfactual invariance in practice. Specifically, we leverage conditional latent diffusion models with disentangled representations to generate high-quality counterfactual images that selectively alter sensitive features while preserving disease-relevant information. This enables the use of well-established statistical tests (e.g., the paired $t$-test for binary classification outputs, or the Kolmogorov--Smirnov test for continuous outputs) to formally assess whether model predictions are invariant to interventions on sensitive attributes. Experimental results on both synthetic and real-world medical imaging datasets, i.e., \textsc{CheXpert} and \textsc{MIMIC-CXR}, demonstrate that this framework adheres closely to the counterfactual invariance principle and outperforms existing baselines.
\section{Related work}
\subsection{Fairness in medical machine learning} 
In recent years, fairness has emerged as an area of focus within the machine learning community. One of the earliest fairness metrics in machine learning is \emph{fairness through unawareness}, which ignores protected attributes like self-reported race and gender. However, this can be ineffective since protected attributes are often encoded in other features \citep{hardt2016equality}. Demographic Parity (DP) refers to the statistical parity of outputs between demographic groups but has been criticized for practically not ensuring fairness \citep{dwork2012fairness, hardt2016equality}, making it unsuitable for medical classification cases. Equality of opportunity (EO) aims for individual fairness but relies on similarities between cases, which are difficult to define in practice \citep{dwork2012fairness, fleisher2021s}. Counterfactual fairness (CF) offers an alternative by ensuring decisions are the same in both real and counterfactual worlds where demographics differ \citep{DBLP:conf/nips/KusnerLRS17, yang2021causal}.

Several works have broadly discussed the application of AI fairness on medical datasets. \citet{Ricci2022FairnessAI} offers a review on fairness in medical imaging but does not propose a novel evaluation method, focusing only on association-based metrics like DP/EO. \citet{Yang2024FairAIMed} investigates the role of demographic encodings and shares a similar fairness goal, but relies on EO, which this work critiques. \citet{Gadgil2024Mechanism} uses explainability tools to analyze predictions of protected attributes but lacks a quantitative bias measure. \citet{Glocker2023ProtectedAttributes} explores encoding and performance disparities but introduces no new methodology, unlike the statistical testing framework proposed in this work. To summarize, existing methods either rely on non-causal metrics or lack a quantified test framework.
\subsection{Counterfactual invariance} Counterfactual invariance broadly refers to the desirable property that a model's predictions remain unchanged when irrelevant aspects of the input are hypothetically altered. Various notions of counterfactual invariance have been proposed in recent years. Almost sure counterfactual invariance means that a variable remains unchanged almost surely, with respect to changes to another variable \citep{fawkes2022selection}. \citet{fawkes2023results,DBLP:conf/nips/VeitchDYE21} study functional counterfactual invariance. According to this definition, a function is considered counterfactually invariant if its output behaves in the same way as in almost sure counterfactual invariance. Distributional counterfactual invariance \citep{fawkes2023results}, on the other hand, requires that the probability distribution of outcomes remains the same under different interventions, rather than the outcomes themselves. Finally, \citet{quizan.2024} studies a generalization of distributional counterfactual invariance, by considering conditional distributions instead. 

A key approach to achieving a predictive model that is counterfactually invariant is learning counterfactual invariant representations \citep{Pogdin.2023, quizan.2024}. \citet{Pogdin.2023} introduces a neural representation that satisfies certain conditional independence conditions. However, their method uses the Hilbert-Schmidt norm of the kernel covariance as penalty during training. This method is a training-time regularizer rather than a post hoc procedure for quantifying invariance in a fixed pretrained model. Hence, it does not apply to the present problem setting. Furthermore, this approach may suffer from high variance, and it may be inapplicable to high-dimensional settings. As for the paper by \citet{quizan.2024}, its method targets counterfactual invariance during the training phase. In contrast, the proposed method quantifies counterfactual invariance in a pretrained model.

Related but methodologically distinct invariance-auditing work is also relevant. PreCOF~\citet{Goethals.2024} uses predictive counterfactual explanations to detect explicit and implicit bias, but is primarily designed for tabular black-box models and does not provide a statistical test of invariance. Vision-based auditing approaches using generated counterfactual images include generative counterfactual face attribute augmentation~\citep{Denton.2019}, attribute-perturbed encoder-decoder network~\citep{Joo.2020}, and Deep-SCM-based approach like~\citep{Dash.2022}. However, these approaches remain distinct from the present setting: \citet{Joo.2020} uses ad hoc slope-based sensitivity analysis and lacks a formal testing pipeline; \citet{Dash.2022} assumes a specified causal graph and structural equations, targets bias scoring and mitigation rather than hypothesis testing. By contrast, the proposed framework is tailored to pretrained medical imaging models and combines high-quality latent-diffusion counterfactuals with standard statistical tests to assess invariance under interventions on sensitive attributes.
\subsection{Diffusion-based generative models} Diffusion-based generative models \citep{sohl-icml-2015, T-DDPM, DBLP:conf/cvpr/RombachBLEO22} have achieved notable success in recent years across a variety of applications. While a comprehensive review is beyond the scope of this work, this paper briefly highlights a few key works relevant to our setting. For a more detailed overview, the authors refer the interested reader to, e.g., \citet{DBLP:journals/tkde/CaoTGXCHL24}. \citet{sohl-icml-2015} introduced diffusion models with a convolutional architecture. \citet{austin-NeurIPS-2021} extended this to discrete diffusion models, achieving competitive results. \citet{ho-NeurIPS-2020} improved the reverse diffusion process by estimating noise, while \citet{nichol-ICML-2021} introduced better noise scheduling for faster sampling. \citet{song-ICLR-2021b} replaced the Markov process with a non-Markovian one, leading to faster sampling. In conditional generation, \citet{dhariwal-NeurIPS-2021} introduces classifier guidance, while \citet{bordes-arXiv-2021} proposes diffusion models that are conditioned on self-supervised representations. \citet{sehwag-CVPR-2022,pandey-NeurIPSW-2021} focus on generating images from low-density regions and using VAEs for controllable image generation, respectively. \citet{ho-arXiv-2021} introduces Cascaded Diffusion Models for high-resolution image generation.
\section{Methods}

\subsection{Overview}
Our approach evaluates whether a medical AI model relies on sensitive attributes, such as the self-reported race or sex, when making predictions. The central idea is to ask a simple question: \emph{Would the model's prediction change if the same patient had a different demographic attribute, while everything else remained identical?} Since such data cannot be observed in reality, we use a generative framework to simulate it. First, each medical image is mapped into a latent representation. Within this representation, we separate (disentangle) the information related to the sensitive attribute from the information that encodes clinical content such as anatomy or disease markers. This separation allows us to adjust the sensitive attribute while keeping all other medical factors unchanged. The generative model then reconstructs a new version of the image from the modified representation. If predictions differ, this indicates that the model is not invariant to the attribute, signaling potential bias. This overview captures the intuition behind our method, while the following sections provide the formal causal definition, model architecture, and statistical testing procedure.

The remainder of this section develops these ideas in detail: Sec. \ref{sec:causality_background} introduces the framework and the notion of invariance that motivates our approach, while Sec. \ref{sec:disentangled_CLDM} presents the disentangled conditional latent diffusion model that enables the generation of realistic alternative images. Finally, we describe how these components are combined into an algorithm for detecting model reliance on sensitive attributes (Sec. \ref{sec:algo}).

\subsection{Background on counterfactual invariance}
\label{sec:causality_background}
We aim to evaluate whether a predictive model relies on a sensitive attribute, such as the self-reported race, when making decisions. Conceptually, this means asking whether the model would give the same prediction for a patient if only their chosen sensitive attribute were different, while all other clinical information stayed the same. To formalize this idea, we adopt the notion of \emph{counterfactual invariance (CI)}, grounded in the framework of probabilistic causality and structural causal models (SCMs, see \citep{pearlj, 10.1214/21-AOS2064}). In this framework, each dataset is viewed as a collection of random variables (r.v.s) generated by a structural equation model, where directed edges represent functional dependencies among variables. The causal structure of the dataset is defined by direct causal effects, i.e., distributional changes induced by \emph{interventions} on the data-generating process (DGP) \citep{10.1214/21-AOS2064}. An intervention amounts to actively manipulating the generative process of a feature in the dataset, without altering the remaining components of the DGP. In this work, we consider perfect interventions $A \gets a$, by which the values of a feature $A$ are set to a constant $A\equiv a$. We use the standard notation $do(A = a)$ to denote such an intervention.

We propose to test the robustness of a ML model using the notion of CI. In our setting, CI is a state in which an ML model's predictions remain unchanged when, in a hypothetical scenario, a specific input variable is altered. As a property of an ML model, CI applies to specific features of the input. For instance, if an ML model is counterfactually invariant to the racial input variable, the prediction for a given input should remain the same regardless of whether the racial variable changes from, say, "Black" to "White". It is important to note that race and racial identity can be difficult attributes to define in practice \citep{Krieger1987}. In this work, we define self-reported race as a social, political, and legal construct, shaped by the interaction between how others perceive an individual and their own self-identification. For our experiments, we rely on patients' self-reported race. Crucially, the $do(A=a)$ operation represents a hypothetical and purely computational manipulation and thus does not imply or advocate altering any real individual's racial identity. Following \citep{DBLP:conf/nips/KusnerLRS17,quizan.2024}, a model $f$ is said to be \textit{counterfactually invariant} with respect to an attribute $A$ if, for any input $X$ with features $\boldsymbol{Z} = \boldsymbol{z}$, its predicted outcome $\hat{y} $ satisfies:
\begin{equation}
\label{eq:count_invariance}
\mathbb{E}[\hat{y} \mid do(A = a), \boldsymbol{Z} = \boldsymbol{z}] = \mathbb{E}[\hat{y} \mid do(A = a'), \boldsymbol{Z} = \boldsymbol{z}]
\end{equation}
for any possible value $a'$ of the attribute $A$. 
\paragraph{Invariance via generative models.}
\label{sec:bridge}

The definition of invariance in Eq.~\eqref{eq:count_invariance} is theoretical, since post-interventional (counterfactual) distributions cannot be directly observed: we never see the same individual under two different demographic attributes. To approximate these unobservable quantities, we rely on \emph{causal generative models}, which provide a principled mechanism for simulating interventions when real counterfactual data are unavailable~\citep{pawlowski2020deep, scholkopf2021toward}. To translate this theoretical notion of counterfactual invariance into a practical image-generation setting, we make the following assumptions:
\begin{enumerate}
    \item \textbf{Data-generating process.} 
    We assume that each observed image $x$ is produced by an underlying generative process that depends on two types of factors: a \emph{sensitive attribute} $a$ (e.g., self-reported race) and a set of \emph{other, independent factors} $\boldsymbol{z}$ (e.g., anatomy, imaging conditions, or noise). Formally, $x = g(a, \boldsymbol{z})$, where $g$ is an unknown function describing how these factors combine to form an image. Changing $a$ while keeping $\boldsymbol{z}$ fixed corresponds to a hypothetical intervention on the attribute while holding all other aspects constant.
    \item \textbf{Latent representation.} 
    We assume the existence of an approximate inverse mapping $h$ that can recover the attribute-invariant factors $\boldsymbol{z}$ from an observed image $x$, i.e., $\boldsymbol{z} \approx h(x)$. This representation $\boldsymbol{z}$ captures the information in the image that is independent of $a$, while discarding information specific to the sensitive attribute.
    \item \textbf{Learnability via conditional diffusion.} 
    We assume that these mappings $g$ and $h$ can be jointly approximated by a conditional latent diffusion model. In particular, the model learns to encode images into attribute-invariant representations $\boldsymbol{z}$ and to generate counterfactual images $\hat{x}_{a'} = g(a', \boldsymbol{z})$ corresponding to new attribute values $a'$.
\end{enumerate}
These assumptions naturally motivate the design of our generative model. In practice, we parameterize these functions using deep neural networks, as described in Sec. \ref{sec:disentangled_CLDM}.

%
%
%
%
%
%
\subsection{Disentangled conditional latent diffusion models}
\label{sec:disentangled_CLDM}
\begin{figure*}[t!]
    \centering
    \includegraphics[width=0.95\linewidth]{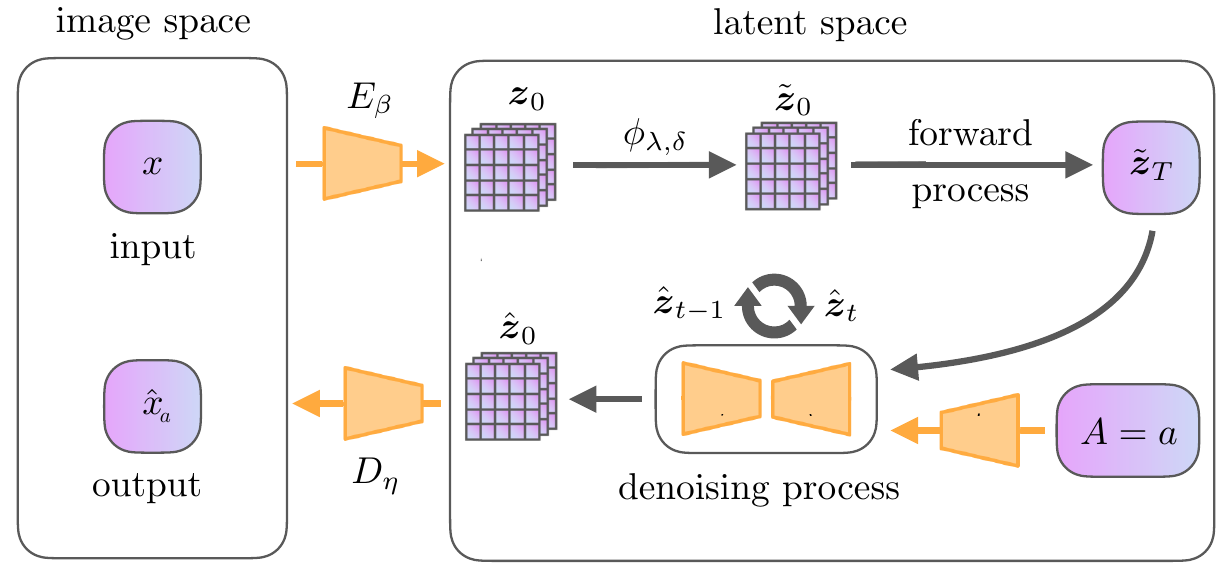}
    \caption{\textbf{Visualization of the disentangled conditional latent diffusion model used to generate images as in Algorithm \ref{alg}.}}
    \label{fig:generative_model}
\end{figure*}
\paragraph{Overview.} 

We introduce the \textbf{Disentangled Conditional Latent Diffusion Model (DCLDM)}, a conditional latent diffusion framework designed to learn latent representations that separate (disentangle) sensitive attributes from other image factors. Intuitively, a DCLDM represents each image in a latent space where the attribute of interest can be modified independently of all other visual characteristics. Similar to latent diffusion models \citep{DBLP:conf/miccai/PinayaTDCFNOC22,DBLP:conf/isbi/PackhauserFTM23,DBLP:conf/cvpr/RombachBLEO22,DBLP:nature_diffusion}, DCLDMs are generative models that perform diffusion processes in a lower-dimensional latent space, enabling efficient and scalable generation of high-quality images \citep{DBLP:conf/miccai/JiangMWCL23}.

\paragraph{Model structure.}
A DCLDM consists of an encoder--decoder pair $(E_\beta, D_\eta)$ that maps images to and from the latent space, a diffusion process $(q_\theta, p_\theta)$ that operates in this space, and a nonlinear transformation $\phi_{\lambda, \delta}$ that enforces disentanglement with respect to the conditioning attribute. The encoder maps an input image $x$ to a latent representation $\boldsymbol{z}_0 = E_\beta(x)$, and the transformation $\tilde{\boldsymbol{z}}_0 = \phi_{\lambda, \delta}(\boldsymbol{z}_0)$ produces an attribute-invariant code. The forward process $q_\theta$ progressively adds noise to $\tilde{\boldsymbol{z}}_0$ over multiple steps, while the reverse process $p_\theta$ learns to denoise these samples conditioned on the attribute value $A = a$. The nonlinear transformation is defined as
\begin{equation*}
\phi_{\lambda, \delta}(\boldsymbol{z}_{0}) = \boldsymbol{z}_{0} + 
\lambda \frac{m_\delta(\boldsymbol{z}_{0})}{\| m_\delta(\boldsymbol{z}_{0})\|},
\end{equation*}
where $m_{\delta}$ is a perturbation network with the same architecture as a U-shaped CNN. This transformation shifts $\boldsymbol{z}_0$ in a direction that reduces its dependence on the sensitive attribute.
\\ $\ $ \\

\noindent \textbf{Why can $\phi_\lambda,\delta$ enforce disentanglement?} The perturbation network $m_{\delta}$ can be interpreted as a learned \emph{attribute-removal direction}: for each latent representation $\boldsymbol{z}_0$, it predicts the direction along which we should shift $\boldsymbol{z}_0$ to make it less informative about the sensitive attribute $A$. This is done by minimizing an approximated mutual information between the representation $\tilde{\boldsymbol{Z}}_0$ and $A$, namely $I(\tilde{\boldsymbol{Z}}_0, A)$.
The normalization in $\phi_{\lambda,\delta}$ ensures that $\lambda$ controls the step size of this shift, so the transformation changes in a controlled way.

\paragraph{Architecture.}
The encoder $E_\beta$ and decoder $D_\eta$ consist of convolutional layers with downsampling and upsampling blocks, respectively, linked by skip connections to preserve spatial information. The diffusion model uses a U-Net architecture~\citep{DBLP:conf/miccai/RonnebergerFB15} for the noise estimator $\epsilon_{\theta}(\boldsymbol{z}_{t}, t, a)$, augmented with cross-attention layers~\citep{DBLP:conf/cvpr/RombachBLEO22} to incorporate the conditioning attribute during denoising.

%
%

\paragraph{Training.} The training of DCLDM involves two phases. In the first phase, the encoder $E_\beta$ and the decoder $D_\eta$ are trained to minimize the standard ELBO loss as in \citet{DBLP:journals/corr/KingmaW13}. In the second phase, the weights of the encoder and decoder are frozen and the images are mapped into the latent vectors using the trained encoder. Then the diffusion model $(p_\theta, q_\theta)$ is trained in the conditional latent space using the conditioned ELBO losses of the diffusion process \citep{DBLP:conf/miccai/PinayaTDCFNOC22,DBLP:conf/isbi/PackhauserFTM23,DBLP:conf/cvpr/RombachBLEO22,DBLP:nature_diffusion}. 

After the model is trained, we train the nonlinear transformation $\phi_{\lambda, \delta} \colon \boldsymbol{z}_0 \to \tilde{\boldsymbol{z}}_0$ to disentangle representations. Following \citet{belghazi2021minemutualinformationneural}, this transformation is learned by minimizing an estimate of the mutual information $I(\tilde{\boldsymbol{Z}}_0; A)$ between the disentangled latent code $\tilde{\boldsymbol{Z}}_0$ and the sensitive attribute $A$. This encourages the transformed representation to be statistically independent of the attribute it was conditioned on (see App.~\ref{app:transformation} for training details).
\paragraph{Generation.} Once the diffusion model is trained in the latent space, it can generate latent representations for a given class label $a$ and an arbitrary random noise $\tilde{\boldsymbol{z}}_T$ as $\hat{\boldsymbol{z}}_{0} \sim p_\theta(\boldsymbol{z} \mid \tilde{\boldsymbol{z}}_T, a)$ using its reverse process. Then, DCLDM generates an image \( \hat{x}\) using the decoder \( D_\eta \), by applying the decoder \( \hat{x}_a = D_\eta(\hat{\boldsymbol{z}}_0) \) to the latent sample \( \hat{\boldsymbol{z}}_0 \). Since the denoised representation \( \hat{\boldsymbol{z}}_0 \) is from the conditional latent space specified by $a$, the reconstructed image also has class label $a$.
\subsection{The proposed algorithm}
\label{sec:algo}
We provide a practical algorithm to test counterfactual invariance from samples using latent representations of a DCLDM. This algorithm is based on the following theorem. 
\begin{theorem}
\label{lemma:cond_variance}
Consider a classifier $f$, e.g., a diagnostic model, with $\{c_1, \dots, c_n \}$ classes. Denote with $\{p_1(x), \dots, p_n(x)\}$ the probability distribution over classes for an input image $x$. Following the notation introduced in Fig~\ref{fig:generative_model}, 
denote with $\hat{x}_a$ the point generated by a DCLDM from a latent representation $\boldsymbol{z}$, conditioned on an attribute $A= a$. Furthermore, denote with $f(\hat{x}_{a})$ the prediction of $f$ on a generated image $\hat{x}_{a}$. Define the quantity
\begin{equation*}
\mathsf{Avg} (\boldsymbol{z}, a) \coloneqq \sum_{i = 1}^C c_i  \cdot p_i (\hat{x}_{a}).
\end{equation*}
Then, the model $f$ is counterfactually invariant with respect to an attribute $A$ if and only if it holds
\begin{equation}
\label{eq:avg}
\mathbb{E} [ \underbrace{ f(\hat{x}_{a}) \cdot \mathsf{Avg} (\boldsymbol{z}, a) }_{\mathsf{LHS}_{\boldsymbol{z},a}}  ]
= 
\mathbb{E} \Big [ \underbrace{ \sum_{\hat{a}\in A}  f(\hat{x}_{\hat{a}}) \cdot \mathsf{Avg} (\boldsymbol{z}, \hat{a}) }_{\mathsf{RHS}_{\boldsymbol{z}}}\Big].
\end{equation}
In this formula, the expected values are taken for varying $a$ and $\boldsymbol{z}$.
\end{theorem}
The proof of this theorem leverages the causal graph structure and the concept of a valid adjustment set to establish conditional independence, which, combined with the law of total expectation and properties of conditional variance, demonstrates the equivalence stated in the theorem (see App.~\ref{sec:lemma_proof} for a detailed proof).
\begin{algorithm}[t]
\footnotesize
\caption{CIT-LR: Conditional Independence Test via Latent Representations}
\label{alg}
\begin{algorithmic}[1]

\STATE \textbf{Input:} Training data $\mathcal{D}_{\text{train}}$, test data $\mathcal{D}_{\text{test}}$, predictive model $f$
\STATE \textbf{Output:} $p$-value testing whether $f$ is counterfactually invariant in $A$.

\vspace{0.5em}
\STATE \textbf{\color{green!60!black}//Step 1: Learn latent representations}
\STATE Train a DCLDM on $\mathcal{D}_{\text{train}}$ with disentangled representations;

\vspace{0.5em}
\STATE \textbf{\color{green!60!black}//Step 2: Estimate test statistics}
\FOR{each test sample $x, a \in \mathcal{D}_{\text{test}}$}
    \STATE Generate counterfactual images $\hat{x}_{\hat{a}}$ for all attributes $\hat{a} \in A$;
    \STATE Compute the quantities $\mathsf{LHS}_{\boldsymbol{z},a} = f(\hat{x}_{a}) \cdot \mathsf{Avg} (\boldsymbol{z}, a) $ and $ \mathsf{RHS}_{\boldsymbol{z}} = \sum_{\hat{a}\in A} f(\hat{x}_{\hat{a}}) \cdot \mathsf{Avg} (\boldsymbol{z}, \hat{a})$
\ENDFOR

\vspace{0.5em}
\STATE \textbf{\color{green!60!black}//Step 3: Perform statistical test}
\STATE Apply a statistical test to determine if $\mathsf{LHS}_{\boldsymbol{z},a}=\mathsf{RHS}_{\boldsymbol{z}}$;

\vspace{0.5em}
\STATE \textbf{return} resulting $p$-value;
\end{algorithmic}
\end{algorithm}

Theorem \ref{lemma:cond_variance} provides a sample-based criterion for evaluating whether a model exhibits counterfactual invariance with respect to an attribute $A$, by linking the model's outputs on generated counterfactuals to a measurable equality involving their expected values. The proposed algorithm is presented in Algorithm \ref{alg}. We refer to this algorithm as Conditional Independence Test via Latent Representations (CIT-LR). Our proposed approach can be summarized in the following steps, based on Theorem \ref{lemma:cond_variance}:
\begin{enumerate}
\item \label{step:1} The algorithm learns a disentangled DCLDM to generate images conditioned on a protected attribute of interest.
\item \label{step:2} Images generated with the learned DCLDM in Step \ref{step:1} are used to learn the inner left-hand side ($\mathsf{LHS}_{\boldsymbol{z},a}$) and right-hand side ($\mathsf{RHS}_{\boldsymbol{z}}$) of Eq. \ref{eq:avg}.
    \item \label{step:3} Use a statistical test to determine if Eq. \ref{eq:avg} holds, by comparing $\mathsf{LHS}_{\boldsymbol{z},a}$ and $\mathsf{RHS}_{\boldsymbol{z}}$.
\end{enumerate}%

\paragraph{Choice of the statistical test in algorithm \ref{alg}.} In our setting, we consider diagnostic models $f$ that have a binary output. For these models, the underlying distributions of $\mathsf{LHS}_{\boldsymbol{z},a}$ and $\mathsf{RHS}_{\boldsymbol{z}}$ are fully determined by their expected values, which correspond to the probability of predicting the positive class. As a result, comparing means in this case is equivalent to comparing distributions. The paired t-test is therefore a natural choice.

We emphasize, however, that in scenarios where the outputs of $f$ are continuous random variables, a full-distribution test such as the Kolmogorov--Smirnov (KS) test would be preferable. The reason is that, unlike in the binary case, the mean no longer uniquely characterizes the distribution. Two continuous distributions may share the same mean yet differ substantially in other aspects such as variance, skewness, or higher-order moments. The KS test is sensitive to any such discrepancy, as it compares the empirical cumulative distribution functions of the two samples and quantifies their maximum deviation. In this way, it is capable of detecting a broader range of distributional shifts, beyond simple mean differences. 

%
%
%
%
%
%
%
%
%
%
%
%
%
%
\subsection{Ethics statement}
This study uses only publicly available and de-identified medical imaging datasets: \textsc{CheXpert} and MIMIC-CXR. The \textsc{CheXpert} dataset is provided by the Stanford Machine Learning Group under the Stanford University Dataset Research Use Agreement. The MIMIC-CXR dataset is available through PhysioNet and is released under the PhysioNet Credentialed Health Data License 1.5.0 and PhysioNet Credentialed Health Data Use Agreement 1.5.0 by the MIT Laboratory for Computational Physiology. Both datasets are de-identified, and the use of these datasets does not require additional Institutional Review Board (IRB) approval. No direct interaction with human participants occurred. This work generates counterfactual images with altered self-reported racial attributes solely for the purpose of studying fairness and mitigating potential model bias, and not for the creation of new identities or raising concerns about individual identity.
\section{Datasets and baselines}
%
%
\subsection{Datasets}
\label{sec:datasets}
\paragraph{Synthetic datasets.} \label{par:syn-data}We consider a synthetic dataset containing 6 diagnostic labels, each designed with a different activation mechanism: \say{linear}, \say{quadratic}, \say{exponential}, \say{interactive}, \say{log-scale}, and \say{sin}. The dataset consists of a sequence of features $Z_1, \dots, Z_n \in \mathbb{R}^{32}$, a protected attribute $A$, a variable $X$ that is causally influenced by both $A$ and $Z_n$, and binary diagnostic labels $Y_1,\ldots,Y_6$. Here, features $Z_1, \dots, Z_n$ are generated using an autoregressive process to create a sequential dependency among them. For reproducibility, we publish the synthetic data on Zenodo (\href{https://doi.org/10.5281/zenodo.18803863}{link)}. 

\paragraph{Medical datasets.} \label{par:real-data} We use two large, publicly available datasets consisting of radiological images, namely the \textsc{CheXpert} dataset \citep{DBLP:conf/aaai/IrvinRKYCCMHBSS19}, and the MIMIC-CXR \citep{mimic-cxr, mimic-cxr-jpg} dataset. The \textsc{CheXpert} dataset is publicly available at \url{https://stanfordaimi.azurewebsites.net/datasets/8cbd9ed4-2eb9-4565-affc-111cf4f7ebe2}. The MIMIC-CXR dataset is publicly available at \url{https://physionet.org/content/mimic-cxr/2.1.0/}. Both datasets consist of chest X-rays paired with radiological reports and demographic attributes of the patients, including the self-reported race. In this work, we restrict the datasets to images with racial labels \say{asian}, \say{black}, and \say{white} to ensure balanced classes and stable training. In this study, we conceptualize self-reported race as a social, political, and legal construct, influenced by both external perceptions and individual self-identification. For the purposes of our experiments, we use patients' self-identified race. For both datasets, images are labeled for five common pathologies: cardiomegaly, edema, consolidation, atelectasis, and pleural effusion, which serve as the classification targets for the models we evaluate. 
\subsection{Baselines} \label{sec:baselines}
First, we compare our proposed method against two most common bias-detection metrics: demographic parity (DP) and equality of opportunity (EO). DP is a fairness metric based on association. According to this metric, a model $f$ is fair if and only if 
\begin{equation}
\label{eq:demographic_parity}
\mathbb{P}[f(X)=1 \mid A=a] = \mathbb{P}[f(X)=1 \mid A=a'].
\end{equation}
DP requires that a model is free of bias if its predictions are statistically independent of the sensitive attribute A. In a diagnostic setting, this means the probability of a positive prediction should be the same across all demographic groups, regardless of the true disease status. To quantify the level of bias under this metric, we perform a paired $t$-test. This choice is well-suited for our setting because $f$ produces binary outputs, where the mean uniquely characterizes the underlying distribution.  
 For instance, when comparing two racial groups (e.g., 'Black' vs. 'White'), we split the model's predictions into two respective demographic subgroups and use the paired $t$-test to determine if their mean prediction rates are significantly different. When multiple subgroups exist, we compute the geometric mean of the pairwise $p$-values.

Equality of opportunity stipulates equal performance for individuals who are actually positively labeled. This means the true positive rate (sensitivity) should be equal across all demographic groups. Formally, a model $f$ satisfies EO w.r.t. the positive outcome iff.
\begin{equation}
\mathbb{P}[f(X)=1 | Y=y, A=a]  = \mathbb{P}[f(X)=1 | Y=y, A=a'] \label{eq:eo}
\end{equation}
Likewise, we transform the criteria into a statistical test, by conducting a paired$t$-test to determine if \eqref{eq:eo} holds. This ensures that DP and EO are both evaluated using the same statistical test, and differences in performance reflect the fairness definitions themselves rather than properties of different tests.

Both DP and EO have been criticized, for either not ensuring fairness \citep{dwork2012fairness, hardt2016equality}, or relying on \say{similarity} attributes that are difficult to define in practice. We remark that DP and EO are purely based on statistical associations, and they do not rely on counterfactuals, as opposed to the notion of counterfactual invariance. In a clinical context, the prevalence of a disease may differ between demographic groups due to socio-economic factors. An accurate, well-calibrated model reflecting these real-world differences could be flagged as "biased" by DP or EO. Furthermore, these metrics can fail to identify subtle, hidden biases that are not reflected in simple statistics.

The baselines DP and EO discussed above are association-based baselines. We further include two counterfactual-generation baselines, which we refer to as CIT-LR (non-disentangled) and ImageCFGen. CIT-LR (non-disentangled) uses the standard CLDM trained identically to Sec.~\ref{sec:disentangled_CLDM} but without the disentangling transformation $\phi_{\lambda,\delta}$. Counterfactual images are generated by conditioning the CLDM on each attribute value $\hat a\in A$, and Algorithm~\ref{alg} is applied unchanged. This baseline serves as an ablation that isolates the effect of disentanglement.

ImageCFGen consists of an adaption of~\citet{Dash.2022} to our setting. For each input pair $(x,a)$ consisting of an image $x$ and an attribute of interest $a$, ImageCFGen produces a reconstructed image $x_i^{r}$ and a counterfactual image $x_i^{cf}$, which corresponds to a modificaton of the protected attribute. To convert this into a counterfactual invariance test, we replace their original descriptive bias score with the paired difference $D_i := f(x_i^{r}) - f(x_i^{cf})$, and test the null hypothesis $H_0:\mathbb{E}[D_i]=0$ using a paired $t$-test for binary outputs. This adaptation is natural because~\citet{Dash.2022} already evaluates a pretrained classifier on matched reconstructed/counterfactual pairs.

\section{Experiments}
We conduct experiments on both synthetic and real-world datasets to compare our framework against DP and EO. The synthetic datasets provide a controlled environment where the ground-truth counterfactual invariance is known, which is essential for validating our method since counterfactual outcomes are unobservable in real-world datasets. In all experiments, Algorithm~\ref{alg} employs a paired $t$-test, since the outputs of $f$ are binary and the mean fully characterizes their distribution. To avoid directly comparing $p$-values, we will use the rejection rate (statistical power) to represent the ability to test model invariance. Formally, Rejection rate measures the probability of a test to reject the null hypothesis. High rejection rates on biased models (low Expected Counterfactual Accuracy) and low rejection rates on unbiased models indicate a high-quality invariance test.  
%

\paragraph{Implementation.} All the code is implemented using Python. For synthetic datasets, the algorithms are built on the statistical learning algorithms in the scikit-learn package. Since these experiments do not require much computational resources, they run on the CPU alone on an Intel Core i7 16GB RAM. For the two real-world medical image datasets, the experiments are run on a single 48GB Nvidia A100 GPU. The code is available at: \url{https://github.com/Neferpitou3871/AI-Alignment-Medical-Imaging}.
\subsection{Alignment on synthetic datasets}
\label{synthetic_data}
In this section, we evaluate our CIT-LR on the synthetic datasets introduced. By using synthetic data, we can compute ground-truth counterfactual labels, which enables a principled and quantitative evaluation of counterfactual invariance. 

\paragraph{Experimental setup.} We construct a pool of $1000$ base classifiers that act as diagnostic AI models. Each classifier takes as input a data point $x \sim X$ and predicts the corresponding label $\hat{y}$. For each classifier, we run Algorithm~\ref{alg} to determine whether it is counterfactually invariant with respect to the attribute $A$. For the data generating process, we consider a scenario where the data $X$ is a mixed interaction of the latent variable $Z$ and protected attribute $A$ while the final label only depends on the latent $Z$. This design allows us to test the ability of the model to disentangle $A$ from input $X$ and thus to assess the counterfactual invariance. Base models, as tested objects, are created from 10 different types of ML models, such as SVM and decision tree. For each model type, we further diversify the hyperparameter configuration and initialize with different random seed to finally create a pool of 1000 base models (see App.~\ref{app:synthetic_classifiers} for details). These models are trained on the synthetic dataset to predict the label $\hat{y}$ based on the input $x$. Once the models' training is completed, we test their CI result using our CIT-LR and the baselines.

\paragraph{Evaluation metric.} We use the expected counterfactual accuracy (ECA) as our evaluation metric. ECA provides a continuous index measuring the extent to which a model satisfies counterfactual invariance. To define ECA, consider the indicator function:
\begin{equation*}
\iota(a, \boldsymbol{z}) := 
\left \{
\begin{array}{ll}
1 & \text{if } \mathbb{E}[\hat{Y} \mid do(A = a), \boldsymbol{Z} = \boldsymbol{z}] = \mathbb{E}[\hat{Y} \mid \boldsymbol{Z} = \boldsymbol{z}]\\
0 & \text{otherwise}
\end{array}
\right .
\end{equation*}
Then, the ECA is given by $\text{ECA} := \mathbb{E}\left [\iota(A, \boldsymbol{Z}) \right ]$. An ECA of $1$ indicates that the intervention has no causal effect beyond what is captured by $\boldsymbol{Z}$, and thus the model $f$ is counterfactually invariant.

\begin{figure*}
    \centering
    \includegraphics[width=1.0\linewidth]{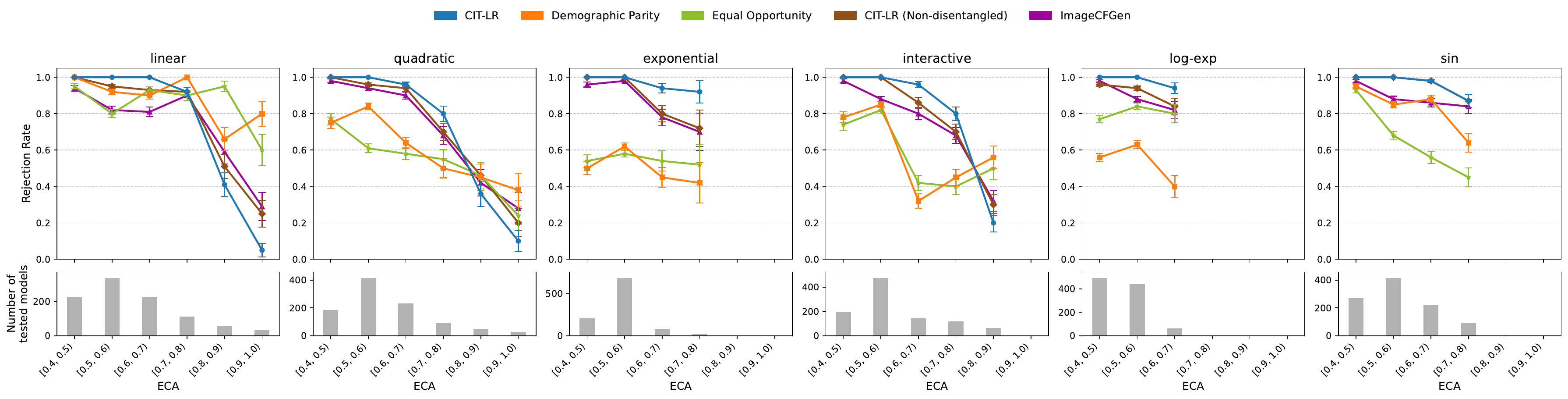}
    \caption{\textbf{Performance comparison of ctf-invariance tests across six tasks on synthetic datasets.} Diagnostic models are grouped into ECA intervals, and for each interval the rejection rates of different methods are displayed. The dot indicates the sample mean rejection rate within the group, while the whiskers represent the sample variance. The number of models within each interval is reported below the corresponding bar.}
    \label{fig:syn_rej_rate}
\end{figure*}

\paragraph{Results.} Fig~\ref{fig:syn_rej_rate} reports the rejection rates of all methods across models grouped by their ECA values, with whiskers indicating variability across repetitions and significance level fixed at $\alpha=0.05$. Overall, CIT-LR exhibits the clearest alignment with the target notion of counterfactual invariance: for biased models ($\mathrm{ECA}<0.8$), it achieves rejection rates close to $100\%$ in nearly all tasks, while DP and EO show significantly weaker and often non-monotone behavior, especially in the exponential, interactive, and log-exp settings. Averaged across all six synthetic tasks, CIT-LR achieves a mean rejection rate of 96.1\% on biased models and a mean false-alarm rate of 22.0\% on near-invariant models ($\mathrm{ECA} \geq 0.8$). The two counterfactual-generation baselines, CIT-LR (Non-disentangled) and ImageCFGen, are consistently stronger than DP and EO, confirming that paired evaluation on generated counterfactuals is more informative than association-based fairness criteria. However, both underperform CIT-LR: when ECA enters the near-invariant regime, CIT-LR reduces its rejection rate much more sharply, reaching the Type~I error level with $\alpha=0.05$ in the linear setting and remaining clearly below the other baselines in the quadratic and interactive tasks, whereas the non-disentangled and ImageCFGen variants still reject at least $25\%$ of the invariant models. This gap is consistent: removing the disentangling leaves sensitive-attribute information in the latent code, while ImageCFGen relies on a specified attribute SCM and ALI-based reconstructed pairs, both of which can reduce counterfactual fidelity and hence weaken the validity of the downstream test. Therefore, the improvement of CIT-LR is not only due to using generated counterfactuals, but specifically due to combining paired hypothesis testing with \emph{disentangled} latent counterfactuals that isolate the sensitive attribute more faithfully.

\subsection{Alignment on real-world datasets}
In this section, we evaluate our CIT-LR on the real-world datasets. 
\paragraph{Experimental setup.} 
We evaluate our method on the \textsc{CheXpert} and MIMIC-CXR datasets, using the same five pathologies and three racial groups (Asian, Black, White) as described in Sec. \ref{sec:datasets}. For each of the real-world datasets, we create $100$ disease classifiers as our tested object for the CIT-LR and the baselines. To create the base classifiers, we train 10 different CNN-based architectures (e.g. DenseNet 121 \citep{densenet} or VGG \citep{vgg}) on each of the image datasets. These base classifiers take as input the image $x \sim X$ and predict the aforementioned five disease labels $\hat{y}$. We sample 10 different checkpoints from each architecture after the validation performance becomes stable (see App.~\ref{app:CheXpert_classifiers} for more details), creating a pool of 100 diagnostic classifiers. For each base classifier, we apply Algorithm \ref{alg} to assess whether it is counterfactually invariant with respect to the attribute $A$. Furthermore, we evaluate DP and EO.
\paragraph{Evaluation metric.} In this case, we cannot compute CI directly, due to the lack of a ground truth. Instead, we compute the expected counterfactual accuracy (ECA). To define this metric, consider the indicator function:
\begin{equation*}
\zeta(a, \boldsymbol{z}) := 
\left \{
\begin{array}{ll}
1 & \text{if } \mathbb{E}[\hat{Y} \mid A=a, \boldsymbol{Z} = \boldsymbol{z}] = \mathbb{E}[\hat{Y} \mid \boldsymbol{Z} = \boldsymbol{z}]\\
0 & \text{otherwise}
\end{array}
\right .
\end{equation*}
Then, the CI is measured by $\text{ECA} := \mathbb{E}\left [\zeta(A, \boldsymbol{Z}) \right ]$. 
We remark that it can be proven that if $\boldsymbol{Z}$ is a disentangled representation of a DCLDM, it then holds $\mathbb{E}[\hat{Y} \mid A=a, \boldsymbol{Z} = \boldsymbol{z}] = \mathbb{E}[\hat{Y} \mid do(A=a), \boldsymbol{Z} = \boldsymbol{z}]$, for any classifier $f$ (see Lemma~\ref{lemma:link_with_diffusion} in App.~\ref{sec:lemma_proof}). Hence, our evaluation strategy offers a reasonable and transparent approximation under current methodological and data availability constraints.
\begin{figure}
    \centering
    \includegraphics[width=0.95\linewidth]{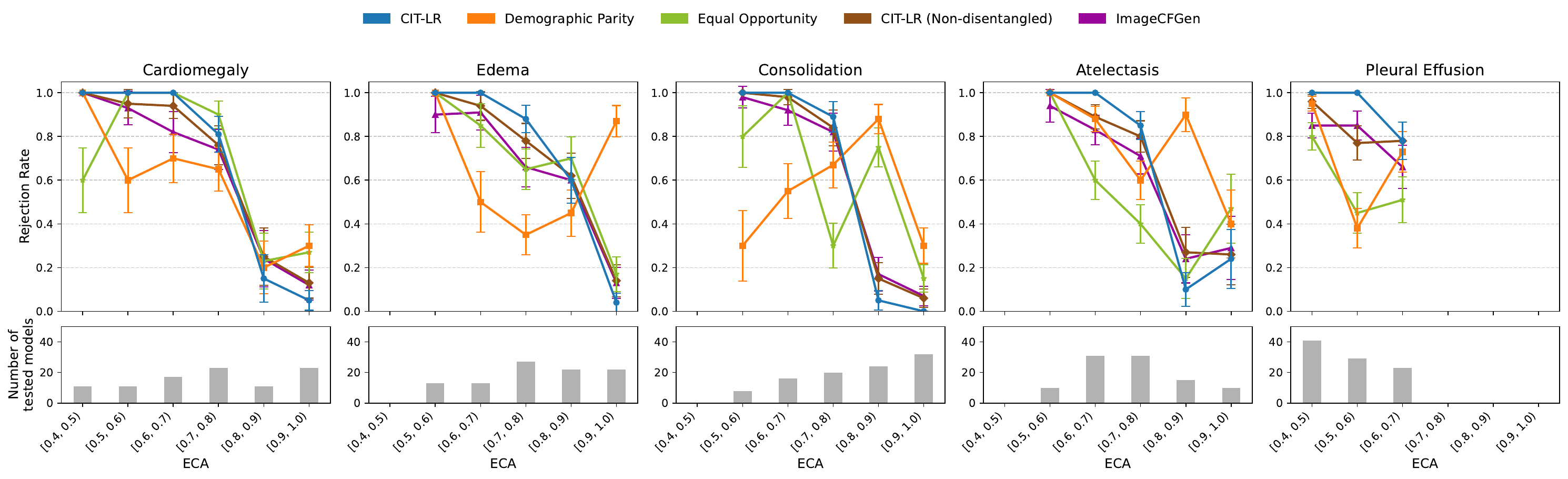}
    \caption{\textbf{Rejection rates of statistical tests on the \textsc{CheXpert} dataset.} Diagnostic models are organized by ECA intervals, and the rejection rates of different methods are shown for each group. The dot represents the sample mean rejection rate within the interval, while the whiskers capture the sample variance. The count of models in each interval is provided below the corresponding bar.}
    \label{fig:cxp_barplot}
\end{figure}

\begin{figure}
    \centering
    \includegraphics[width=0.95\linewidth]{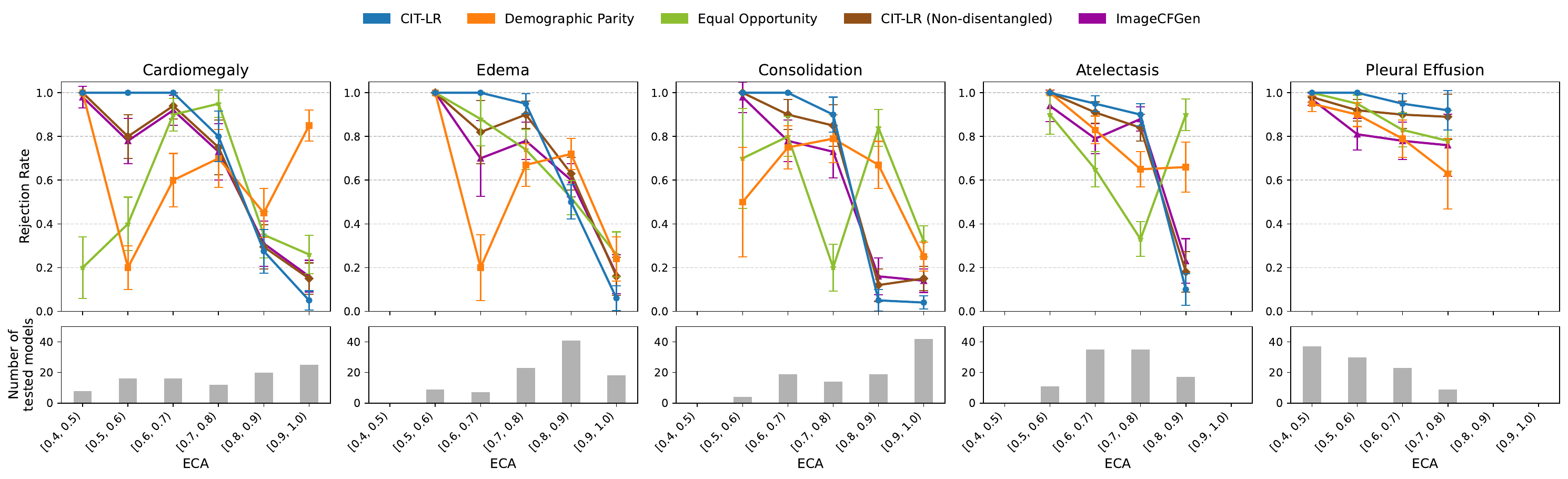}
    \caption{\textbf{Rejection rates of statistical tests on the \textsc{MIMIC-CXR} dataset.} Diagnostic models are organized by ECA intervals, and the rejection rates of different methods are shown for each group. The dot represents the sample mean rejection rate within the interval, while the whiskers capture the sample variance. The count of models in each interval is provided below the corresponding bar.}
    \label{fig:mimic_barplot}
\end{figure}
\paragraph{Results.} Figs~\ref{fig:cxp_barplot}--\ref{fig:mimic_barplot} illustrate the rejection rates of various statistical tests across models grouped by their ECA values, for \textsc{CheXpert} and \textsc{MIMIC-CXR} respectively. Each plot shows the average rejection rate, with whiskers representing variability across repetitions. For all experiments, the significance level is fixed at $\alpha = 0.05$. In both datasets, CIT-LR shows the strongest alignment with ECA: for low-ECA models ($\text{ECA} < 0.8$) it rejects at high rates close to $100\%$ for Cardiomegaly, Edema, Consolidation, and Atelectasis; then decreases substantially once the models move into higher-ECA bins ($\text{ECA} \in [0.8, 1.0)$). This drop is particularly clear on \textsc{MIMIC}, where for Cardiomegaly, Edema, Consolidation, and Atelectasis the CIT-LR rejection rate falls to about $0\%$--$15\%$ in the highest available ECA bins; on \textsc{CheXpert} the same downward trend is visible, although some tasks such as Cardiomegaly and Atelectasis remain slightly above zero in the last bin. Averaged across all five diagnostic tasks, CIT-LR achieves a mean rejection rate of 96.3\% on biased models for CheXpert and 95.7\% for MIMIC-CXR, while the mean false-alarm rate drops to 15.3\% and 14.7\%, respectively, for the least biased models ($\mathrm{ECA} \geq 0.8$). In contrast, DP and EO are more irregular across bins: both often have much lower rejection rates on clearly biased models ($\text{ECA} \in [0.5, 0.6)$) and several non-monotone increases in higher-ECA bins, as seen for example in Edema, Consolidation, and Atelectasis. The two counterfactual-generation baselines, CIT-LR (Non-disentangled) and ImageCFGen, are generally closer to CIT-LR than DP and EO in the low-ECA regime, but in the higher-ECA bins they usually remain above CIT-LR, with visibly slower declines for tasks such as Cardiomegaly and Edema and, in \textsc{MIMIC} Consolidation, rejection rates still around $10\%$--$20\%$ when CIT-LR is already near zero. For Pleural Effusion, the figure contains only lower-ECA models in both datasets, and all counterfactual-based methods stay at comparatively high rejection rates, with our CIT-LR achieving the highest rejection rates consistently.

\subsection{Generative performance of DCLDM}
\label{sec:dlcdm-image}
\begin{figure}[t]
    \centering
    \includegraphics[width=1\linewidth]{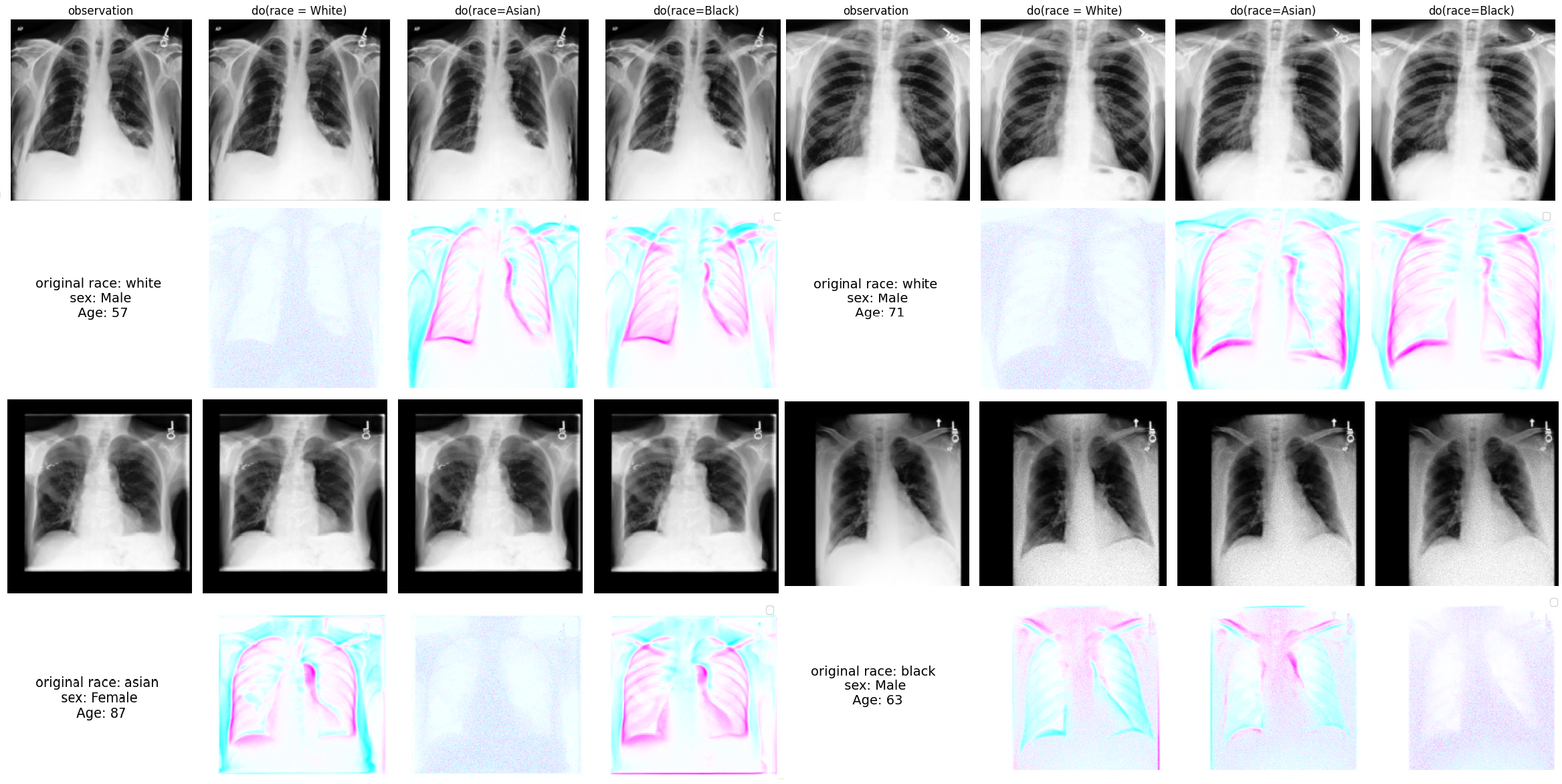}
    \caption{\textbf{Generated images using our DCLDM.} The first and third rows show original and counterfactual images from \textsc{CheXpert} and \textsc{MIMIC}, respectively. The second and fourth rows show the corresponding pixel-wise differences between the original and generated images.}
    \label{fig:ctf-img}
\end{figure}

%
%
%
%
The analysis on real-world imaging datasets relies on images generated with our DCLDM (see Fig~\ref{fig:ctf-img}). Therefore, we conduct an additional set of experiments to validate its generative performance. We focus on two key aspects: (1) The reconstructed image preserves the essential anatomical and pathological features of the original image; (2) the detected racial attribute of the reconstructed image is consistent with the counterfactual label.

\paragraph{Experimental setup.} We take two steps to build the DCLDM: Step 1: We train a standard CLDM, which consists of an encoder, a decoder and an inner latent diffusion model, on the two image datasets. Step 2: We train the disentangling transformation that creates the counterfactual latent representation (see App.~\ref{app:cldm} and App.~\ref{app:transformation} for implementation details). 

\paragraph{Evaluation metric.}
We evaluate our model and baselines using two complementary metrics. First, Effectiveness Accuracy measures how well the model achieves the desired counterfactual attribute. Specifically, it is defined as the proportion of generated images whose predicted racial label matches the conditioning label provided during generation. For this evaluation, we employ a high-performance ResNet-101 (see App.~\ref{app:race_classifier}) classifier pre-trained to detect racial attributes from images \citep{Banerjee2021AIRO}. Second, the Fr\'echet Inception Distance (FID) assesses the overall quality of the generated images by comparing the statistical properties of their features to those of real images, with lower FID scores indicating higher realism. Fig~\ref{fig:ctf-img} illustrates representative samples generated by our DCLDM alongside their original counterparts.
\begin{table}[t]
\centering
\caption{\textbf{Effectiveness accuracy and FID on the \textsc{CheXpert} and MIMIC-CXR test datasets.} }
\label{tab:evalctf2}
\begin{tabular*}{\textwidth}{@{\extracolsep{\fill}}lcccc}
\toprule
\multirow{2}{*}{\textbf{Dataset}} & \multicolumn{3}{c}{\textbf{Effectiveness Accuracy (\%) $\uparrow$}} & \multirow{2}{*}{\textbf{FID $\downarrow$}} \\
\cmidrule(lr){2-4}
 & \textit{A = \emph{Black}} & \textit{A = \emph{Asian}} & \textit{A = \emph{White}} & \\
\midrule
\textsc{CheXpert}   & 88.7 & 91.8 & 93.3 & 21.79 \\
\textsc{MIMIC-CXR}  & 89.2 & 92.0 & 93.7 & 25.89 \\
\bottomrule
\end{tabular*}
\end{table}
\paragraph{Results.} The quantitative results of our validation experiments are displayed in Table \ref{tab:evalctf2}. Across both the \textsc{CheXpert} and MIMIC-CXR datasets, our DCLDM demonstrates good performance. It achieves high Effectiveness Accuracy, indicating that the desired conditional attribute is embedded correctly. Furthermore, our DCLDM obtains low FID scores, indicating the generation of high-quality and realistic images.
%
%

\subsection{Validating the absence of artifacts in image generation}

A central requirement for validating our generative framework is to ensure that the synthesis process does not introduce spurious artifacts or detectable digital fingerprints. If such irregularities were present, they could bias downstream diagnostic models toward exploiting unintended visual cues rather than clinically meaningful content. To address this concern, we design experiments that explicitly evaluate whether generated images are visually indistinguishable from authentic ones, thereby testing the fidelity of the generative process.

\paragraph{Experimental set-up.}  
For each pair of source and target attributes in the dataset, where attributes correspond to self-reported race $\{\text{Black, White, Asian}\}$, we define a balanced binary classification task. A ResNet-101 model (pre-trained on ImageNet) is fine-tuned to distinguish real target-attribute images from synthetic target-attribute images generated by converting from the source attribute. Real images are assigned label $0$, and synthetic images label $1$. To avoid disease-related confounding, we restrict the analysis to healthy patients only. Each task uses $2500$ real and $2500$ synthetic images, which are randomly partitioned into training ($80\%$), validation ($10\%$), and test ($10\%$) sets.

\paragraph{Evaluation metric.}  
We assess the presence of artifacts using an adversarial evaluation strategy. For the ResNet-101 models trained as above, we measure performance using the AUC-ROC metric. Chance-level AUC-ROC scores indicate that the classifier cannot distinguish between the two sets of images, providing strong evidence that the generative process does not introduce detectable artifacts.

\begin{figure}[ht]
    \centering
    \includegraphics[width=\linewidth]{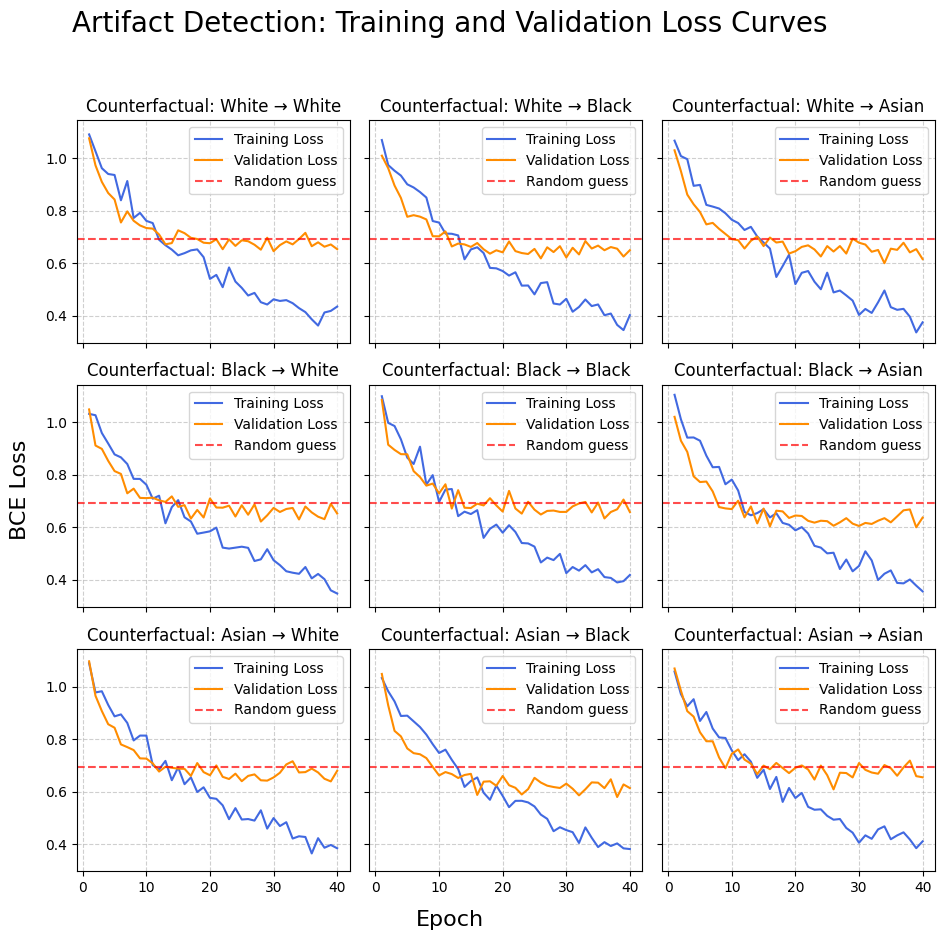}
    \caption{\textbf{Training and validation loss curves for pairwise real-vs-synthetic classification.} The curves show stable convergence with no evidence of overfitting.}
    \label{fig:train-val-loss}
\end{figure}

\paragraph{Results.}  
The pairwise AUC-ROC scores on the held-out test set are summarized in Table~\ref{tab:artifact_results_full}. Across diagonal entries, where source and target attributes coincide, the AUC values are consistently near $0.51$--$0.52$, confirming chance-level performance and indicating that synthetic images are indistinguishable from real ones. Off-diagonal entries show only modest increases in AUC, which we attribute to residual features from attribute conversion rather than artifact introduction. Fig~\ref{fig:train-val-loss} further supports these findings, showing stable training and validation loss curves with no evidence of overfitting. Consistent with the near-random test performance, the classifiers fail to generalize, reinforcing the conclusion that the generative model does not introduce systematic artifacts.
\begin{table}[t]
\centering
\caption{\textbf{Artifact detection results (AUC-ROC).} Diagonal entries correspond to identity transformations, while off-diagonal entries represent cross-race conversions. Each cell reports the performance of a ResNet-101 classifier trained to distinguish real target-race images (columns) from synthetic images generated from a source race (rows). AUC values near $0.5$ indicate that real and synthetic images are indistinguishable.}
\label{tab:artifact_results_full}
\begin{tabular*}{\textwidth}{@{\extracolsep{\fill}}lccc}
\toprule
\multirow{2}{*}{\textbf{Source Race}} & \multicolumn{3}{c}{\textbf{Target Race}} \\
\cmidrule(l){2-4}
 & White & Black & Asian \\
\midrule
\textbf{White} & 0.51 $\pm$ 0.02 & 0.56 $\pm$ 0.04 & 0.55 $\pm$  0.03 \\
\textbf{Black} & 0.54 $\pm$ 0.03 & 0.52 $\pm$ 0.02 & 0.57 $\pm$ 0.02 \\
\textbf{Asian} & 0.55 $\pm$ 0.02 & 0.58 $\pm$ 0.03 & 0.51 $\pm$ 0.03 \\
\bottomrule
\end{tabular*}
\end{table}

\subsection{Effectiveness of signal removal for the sensitive attribute}
\label{sec:signal-removal}

In addition to verifying that our method does not introduce artifacts, we conduct a further experiment to evaluate an equally important property: the removal of signals associated with self-reported race during the generative process. The motivation behind this experiment is to determine whether our counterfactual generator can effectively eliminate identifiable demographic information from images, thereby reducing the ability of downstream models to distinguish between self-identified racial groups after conversion.

\paragraph{Experimental set-up.} The experiment is carried out in two steps. (1)~We first train a ResNet-101 classifier to distinguish between pairs of self-identified racial groups in real images. For each pair of groups, we construct a balanced dataset of $5000$ images, with $2500$ images from each group, and split it into training, validation, and test subsets in an $80\%, 10\%, 10\%$ ratio. We then fine-tune a ResNet-101 classifier (pre-trained on ImageNet) on this dataset to separate the two groups. (2)~Next, we apply our counterfactual generator to transform images from one group so that their demographic label matches that of the other group. The trained classifier is then evaluated, without retraining, on a combined test set containing real images from the other group and the transformed images. This procedure allows us to measure how much the classifier's performance degrades once demographic signals tied to self-reported race are reduced.

\paragraph{Evaluation metrics.} To quantify the extent of signal removal, we use the area under the ROC curve (AUC) as the primary evaluation metric of the ResNet-101 model trained as described above. A high baseline AUC indicates that real images from different self-identified racial groups are easily separable by the classifier. After applying the counterfactual conversion, a substantial reduction in AUC would suggest that demographic cues linked to self-reported race have been effectively reduced, since the classifier is no longer able to distinguish between groups.

\paragraph{Results.} The outcomes of this experiment are summarized in Table~\ref{tab:signal_removal_results}. The "Baseline AUC" column shows consistently high values (above $0.96$), confirming that real images contain strong demographic signals related to self-reported race, which the trained classifier can reliably detect. By contrast, the "Post-Conversion AUC" column demonstrates a significant performance drop when the classifier is evaluated on counterfactual images, providing strong evidence that our generator successfully reduces identifiable signals of self-reported race.

\begin{table}[t]
\centering
\caption{\textbf{Performance of classifiers before and after conversion.} High Baseline AUC values indicate that real images contain identifiable signals of self-reported race, while Post-Conversion AUC values near 0.5 demonstrate that these signals are substantially reduced by our method.}
\label{tab:signal_removal_results}
\begin{tabular*}{\textwidth}{@{\extracolsep{\fill}}lcc}
\toprule
\textbf{Classifier Task} & \textbf{Baseline AUC} & \textbf{Post-Conversion AUC} \\
\cmidrule(r){2-2} \cmidrule(l){3-3}
 & (Real vs. Real) & (Real vs. Synthetic) \\
\midrule
White vs. Black & 0.98 $\pm$ 0.01 & 0.56 $\pm$ 0.03 \\
White vs. Asian & 0.97 $\pm$ 0.02 & 0.59 $\pm$ 0.04 \\
Black vs. Asian & 0.97 $\pm$ 0.02 & 0.58 $\pm$ 0.03 \\
Black vs. White & 0.98 $\pm$ 0.03 & 0.55 $\pm$ 0.03 \\
Asian vs. White & 0.96 $\pm$ 0.04 & 0.57 $\pm$ 0.05 \\
Asian vs. Black & 0.97 $\pm$ 0.03 & 0.56 $\pm$ 0.04 \\
\bottomrule
\end{tabular*}
\end{table}

\subsection{Disentanglement on latent space}
\label{sec:disentangle}
In Section~\ref{sec:signal-removal}, we show that the racial signal is reduced in the \emph{image space}. Here, we conduct an additional experiment to assess this assumption directly in the \emph{latent space}, which is central to our method. The motivation is to test whether the proposed nonlinear transformation $\phi_{\lambda,\delta}$ reduces the recoverability of self-reported race from the latent representation itself.

\paragraph{Experimental setup.}
For each image $x$, we obtain both the original latent representation $z_0 = E_\beta(x)$ and the transformed latent representation $\tilde z_0 = \phi_{\lambda,\delta}(z_0)$. Then, we freeze the encoder $E_\beta$ and $\phi_{\lambda,\delta}$, and train probes to predict the self-reported race label from original and transformed latent vectors respectively. We use two kinds of probes: (i) a linear probe, instantiated with a logistic classifier, and (ii) a small nonlinear probe, instantiated with a two-layer MLP with ReLU activations and early stopping. This yields four probe settings in total: linear probe on $z_0$, linear probe on $\tilde z_0$, MLP probe on $z_0$, and MLP probe on $\tilde z_0$. The experiment is conducted separately on \textsc{CheXpert} and MIMIC-CXR. From each dataset, we randomly sample 3000 images, including 1000 white, 1000 Asian, and 1000 black images. We split the data into training, validation, and test subsets in an $80\%, 10\%, 10\%$ ratio. The results are averaged across multiple random seeds.

\paragraph{Evaluation metrics.}
We report the one-vs-rest AUC on the held-out test set. Lower AUC on $\tilde z_0$ indicates that racial information is less recoverable after the transformation.

\paragraph{Results.}
The AUC results are reported in Table~\ref{tab:latent-probe}. For both datasets and both probe instantiations, the pre-transformation AUC shows that the latent $\boldsymbol{z}_0$ contains rich information about race, with mean AUC above 0.79. In contrast, the AUC scores drop to near chance levels on the transformed representations. For example, on \textsc{CheXpert}, the MLP probe macro-AUC decreases from 0.89 on $z_0$ to 0.52 on $\tilde z_0$, while the linear probe decreases from 0.83 to 0.50. This confirms that the racial signal exists in the original latent vector but is removed after the transformation. 

\begin{table}[t]
\centering
\caption{Results of latent-space probing experiments. AUC (macro-averaged) measures how well an MLP can predict self-reported race from latent codes.}
\label{tab:latent-probe}
\begin{tabular*}{\textwidth}{@{\extracolsep{\fill}}lccc}
\toprule
\textbf{Dataset} & \textbf{Probe} & \textbf{Baseline AUC} & \textbf{Post-transformation AUC} \\
\cmidrule(r){3-3} \cmidrule(l){4-4}
 &  & \textbf{Original Latent $\boldsymbol{z}_0$} & \textbf{Disentangled Latent $\tilde{\boldsymbol{z}}_0=\phi_{\lambda,\delta}(z_0)$} \\
\midrule
\textsc{CheXpert}  & MLP & 0.89 $\pm$ 0.02 & 0.52 $\pm$ 0.04  \\
\textsc{CheXpert}  & Linear & 0.83 $\pm$ 0.03 & 0.50 $\pm$ 0.04  \\
\textsc{MIMIC-CXR} & MLP & 0.82 $\pm$ 0.03 & 0.55 $\pm$ 0.03 \\
\textsc{MIMIC-CXR} & Linear & 0.79 $\pm$ 0.02 & 0.51 $\pm$ 0.03 \\
\bottomrule
\end{tabular*}
\end{table}

\section{Discussion}
The proposed counterfactual invariance test via latent representations has significant implications for fairness in medical AI. The method explores the deep causal mechanism of the diagnostic models, revealing dependencies, which are overlooked by traditional association-based metrics. Through quantifying the biases linked to the sensitive attribute, our method identifies the diagnostic models' undesirable dependencies across demographic groups, reducing the risk of misdiagnosis. Overall, this work advances responsible AI in healthcare, encouraging further improvement of AI fairness. 

However, challenges are also present. For instance, the hard-to-control variance and complexity of diffusion models make the generative model hard to train. In addition to model complexity, computational efficiency is another practical consideration. Although our experiments were conducted on accessible hardware---a single CPU (Intel Core i7, 16~GB RAM) and one NVIDIA~A100~GPU (48~GB)---future work could explore more lightweight architectures or apply model-distillation techniques to further reduce computational demands and enhance accessibility across institutions.

In the experiment, we rely on this proxy measure ECA to assess alignment of the test framework, which, while informative, is inherently limited in its ability to fully capture true causal effects. We acknowledge that this introduces a degree of uncertainty in our evaluation. However, this approach reflects a broader constraint in the field, where ethical, logistical, and practical considerations prevent direct observation of counterfactuals.

The validity of the CIT-LR relies on the DCLDM's ability to satisfy the causal disentanglement assumption. Failure modes might occur if 1) the sensitive information is not fully disentangled, or 2) the decoder introduces anatomical artifacts which get perceived as disease markers by the diagnostic model. In such cases, the test might reject a fair model or accept an unfair model. For this reason, diagnostics on the DCLDM (e.g., real-vs-synthetic AUC near 0.5, stable FID across groups, and qualitative inspection by domain experts) should be conducted to ensure the correct functioning of the method.

The present study focuses on auditing a single sensitive attribute. In principle, the framework can be extended easily to joint attributes $A=(A_1,\ldots,A_k)$ by conditioning the generator on multi-attribute labels. However, users should be careful about interactions within attributes. For example, race and sex might jointly bias the diagnostic, even if single-attribute tests appear insignificant. Furthermore, the assumption that certain attributes should be considered "protected" requires careful consideration. Consultation with medical professionals is advised to build an accurate causal understanding. Moreover, we advise practical validation by clinical experts, such as radiologists, to examine the generated counterfactual images to further confirm the anatomical correctness. 

Future work should aim to refine these representations and address unobserved confounding and expand the methodology to cover a broader range of problems. Although our study focused on self-reported race, it is interesting to extend this framework to other factors such as hospital location or imaging site. Because imaging devices and acquisition protocols vary across institutions, exploring these sources of variation represents an important next step. Doing so could enable fairness evaluation across diverse clinical environments and lead to models that generalize more effectively across healthcare systems. More broadly, the framework is modality-agnostic: any setting where a conditional generator can approximate counterfactual image generation admits the use of our CIT-LR. Extending beyond chest X-rays to CT/MRI/ultrasound or dermoscopy would primarily require a modality-specific CLDM, e.g., a 3D CLDM for brain imaging. This highlights the broader applicability of our approach and opens a clear pathway for future work across diverse medical imaging settings.

\section{Acknowledgments}
The authors gratefully acknowledge the Gauss Centre for Supercomputing e.V. (www.gauss-centre.eu) for providing computing time through the John von Neumann Institute for Computing (NIC) on the GCS Supercomputer JUPITER | JUWELS at J{\"u}lich Supercomputing Centre (JSC).

\bibliographystyle{apalike}
\bibliography{references}
\newpage
\renewcommand{\thesection}{\Alph{section}}
\setcounter{section}{0}
\noindent {\LARGE\textbf{Appendix}}
\section{Missing proofs} \label{sec:lemma_proof}
\subsection{Background on structural causal models}
\label{appendix:scm}
\begin{definition}[Structural Causal Model (SCM), Definition 2.1 by \citet{10.1214/21-AOS2064}]
\label{def:SCM}
A structural causal model (SCM) is a tuple $\langle \mathtt{I},\mathtt{J}, \boldsymbol{V}, \boldsymbol{U}, \boldsymbol{f}, \mathbb{P}_{\boldsymbol{U}} \rangle $ where
(i) $\mathtt{I}$ is a finite index set of endogenous variables; (ii) $\mathtt{J}$ is a disjoint finite index set of exogenous variables; (iii) $\boldsymbol{V} = \prod_{j \in \mathtt{I}} V_j$ is the product of the domains of the endogenous variables, where each $V_j$ is a standard measurable space; (iv) $\boldsymbol{U} = \prod_{j \in \mathtt{J}} U_j$ is the product of the domains of the exogenous variables, where each $U_j$ is a standard measurable space; (v) $\boldsymbol{f} \colon \boldsymbol{V} \times \boldsymbol{U} \to \boldsymbol{V}$ is a measurable function that specifies the causal mechanism; (vi) $\mathbb{P}_{\boldsymbol{U}} = \prod_{j \in \mathtt{J}} \mathbb{P}_{\boldsymbol{U}_j}$ is a product measure, where $\mathbb{P}_{\boldsymbol{U}_j}$ is a probability measure on $\boldsymbol{U}_j$ for each $j \in \mathtt{J}$.
\end{definition}
In SCMs, the functional relationships between variables are expressed in terms of deterministic equations. This feature allows us to model the cause-effect relationships of the data-generating process (DGP) using \emph{structural equations}. For a given SCM $\langle \mathtt{I},\mathtt{J}, \boldsymbol{V}, \boldsymbol{U}, \boldsymbol{f}, \mathbb{P}_{\boldsymbol{U}} \rangle $ a structural equation specifies an endogenous random variable $V_l$ via a measurable function of the form $V_l = f_{V_l}(\boldsymbol{V}, \boldsymbol{U})$ for all $l \in \mathtt{I}$. A \emph{parent} $i \in \mathtt{I} \cup \mathtt{J}$ of $l$ is any index for which there is no measurable function $g \colon \prod_{j \in \mathtt{I}\setminus \{i\}} V_j \times \boldsymbol{U} \to V_l$ with $f_{V_l} = g$ almost surely. Intuitively, each endogenous variables $V_j$ is specified by its parents together with the exogenous variables, via the structural equations. A structural equations model as in Definition \ref{def:SCM} can be conveniently described with the \emph{causal graph}, a directed graph of the form $\mathcal{G} = (\mathtt{I}\cup \mathtt{J}, \mathcal{E})$. The nodes of the causal graph consist of the entire set of indices for the variables, and the edges are specified by the structural equations, i.e., $\{ j \to l \} \in \mathcal{E}$ iff $j$ is a parent of $l$. Note that the variables in the set $\pa{V_l}$ are indexed by the parent nodes of $l$ in the corresponding graph $\mathcal{G}$.%

\paragraph{Interventions.}
We define the causal semantics of SCMs, by considering perfect interventions \citep{pearlj}. For a given a SCM as in Definition \ref{def:SCM}, consider a variable $ \boldsymbol{W} \coloneqq \prod_{j \in \mathtt{I}'}V_j$ for a set $\mathtt{I}' \subseteq \mathtt{I}$, and let $\boldsymbol{w}\coloneqq \prod_{j \in \mathtt{I}'}v_j$ be a point of its domain. The perfect intervention $\boldsymbol{W} \gets \boldsymbol{w}$ amounts to replacing the structural equations $V_{j} = f_{V_{j}}(\boldsymbol{V}, \boldsymbol{U})$ with the constant functions $V_{j} \equiv v_{j}$ for all $j \in \mathtt{I}'$. We denote with $V_l \mid do(\boldsymbol{w})$ the variable $V_l$ after performing the intervention. This procedure define a new probability distribution $\pr{}{v_l \mid do(\boldsymbol{W} = \boldsymbol{w})}$, which we refer to as interventional distribution. This distribution entails the following information: “Given that we have observed
$\boldsymbol{W} = \boldsymbol{w}$, what would $V_l$ have been had we set $do(\boldsymbol{W} = \boldsymbol{w})$, instead of the value $\boldsymbol{W}$ had actually taken?”.
\subsection{Proof of Theorem \ref{lemma:cond_variance}}
In order to prove Theorem \ref{lemma:cond_variance}, we use the well-known concept of valid adjustment set in causal inference. For a review on this notion and its properties we refer the reader to , e.g., \citet{causality,elementscausality}.\\

\begin{definition}[Valid Adjustment Set]
\label{def:valid_adjustment_set}
Let $\mathcal{G}$ be a causal graph and let $A$, $\hat{Y}$ be disjoint (sets of) nodes in $\mathcal{G}$. A set of nodes $\boldsymbol{W}$ is a valid adjustment set for $\{A, \hat{Y}\}$, if
(i) no element in $\boldsymbol{W}$ is a descendant of any node which lies on a proper causal path from $A$ to $\hat{Y}$, in the graph obtained by removing from $\mathcal{G}$ all incoming arrows into nodes in $A$; (ii) $\boldsymbol{W}$ blocks all non-causal paths from $A$ to $\hat{Y}$ in $\mathcal{G}$.
\end{definition}
Valid adjustment sets have the following important property: Let $\boldsymbol{W}$ be a valid adjustment set for $\{A, \hat{Y}\}$. Then, it holds $ \pr{}{\hat{Y} \mid do(A = a) } = \sum_{w} \pr{}{\hat{Y} \mid A = a, \boldsymbol{W} = \boldsymbol{w} } \pr{}{\boldsymbol{W} = \boldsymbol{w}}$, for all values $a$ in the support of $A$. We now have all the tools to prove the following lemma.\\

\begin{figure*}[t]
    \centering
 \includegraphics[height=4cm]{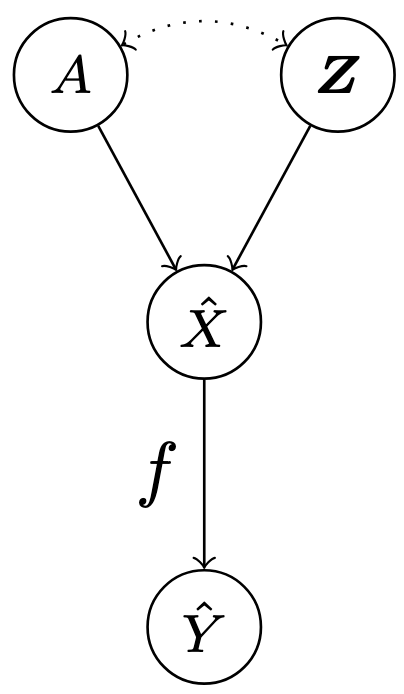}
  \caption{%
    \textbf{Causal structure for the DGP induced by the DCLDM.}
}
\label{fig:causal_graph_gen}
\end{figure*}

\begin{lemma}
\label{lemma:link_with_diffusion}
Denote with $\hat{X}_a$ the random variable of a DCLDM and let $\hat{X}_A$ be the random variable obtained by evaluating the conditional family $\{\hat{X}_a\}_{a\in A}$ at the random variable $A$. Consider the predicted outcome $\hat{Y} = f(\hat{X}_A)$, for a classifier $f$. Then, it holds $\mathbb{P}(\hat{Y} \mid do(A = a), \boldsymbol{Z} = \boldsymbol{z}) = \mathbb{P}(\hat{Y} \mid A = a, \boldsymbol{Z} = \boldsymbol{z})$, for all possible values of $A$.
\end{lemma}
\begin{proof} Consider the generative process given by the DCLDM. In order to generate an image $\hat{x}_a$, we sample a representation vector $\boldsymbol{z}_T\sim \boldsymbol{Z}$ according to a Gaussian prior, and denoise this representation vector using a conditioning label $a\sim A$. Then, the predicted label of interest $\hat{Y}$ is given as $\hat{y} = f(\hat{x}_a)$. This generative process can be visualized with a causal graph as in Fig. \ref{fig:causal_graph_gen}. 
In this graph, the empty set is a valid adjustment set for $\{A,\hat{Y}\}$. Hence, the claim follows.
\end{proof}
We can now prove the main result, which we restate for completeness.
\begin{theorem}[Restatement of Theorem \ref{lemma:cond_variance}]
Consider a classifier $f$, e.g., a diagnostic model, with $\{c_1, \dots, c_n \}$ classes. Denote with $\{p_1(x), \dots, p_n(x)\}$ the probability distribution over classes for an input image $x$. For an input image $x$, denote with $f(\hat{x}_{a})$ the prediction of $f$ on a generated image with conditioning label $A = a$. Furthermore, define the quantity
\begin{equation*}
\mathsf{Avg} (\boldsymbol{z}, a) \coloneqq \sum_{i = 1}^C c_i  \cdot p_i (\hat{x}_{a}),
\end{equation*}
where $\hat{x}_{a}$ is sampled by denoising the representation $\boldsymbol{z}$ conditioned on $a$. Then, the model $f$ is counterfactually invariant with respect to an attribute $A$ if and only if it holds
\begin{equation}
\label{eq:avg}
\mathbb{E} \left [ f(\hat{x}_{a}) \cdot \mathsf{Avg} (\boldsymbol{z}, a) \right ] = \sum_{\hat{a}\in A} \mathbb{E} \left [ f(\hat{x}_{\hat{a}}) \cdot \mathsf{Avg} (\boldsymbol{z}, \hat{a}) \right ]
\end{equation}
In this formula, the expected values are taken for varying $a$ and $\boldsymbol{z}$.
\end{theorem}
\begin{proof}
Following the notation of Lemma \ref{lemma:link_with_diffusion}, note that by the conditional ignorability assumption we have that $\mathbb{E}[\hat{Y} \mid do(A = a), \boldsymbol{Z} = \boldsymbol{z}] = \mathbb{E}[\hat{Y} \mid A = a, \boldsymbol{Z} = \boldsymbol{z}]$ for all $a$. Hence, a model $f$ is counterfactually invariant with respect to the attribute $A$ if it holds $\mathbb{E}[\hat{Y} \mid A= a, \boldsymbol{Z} = \boldsymbol{z}] = \mathbb{E}[\hat{Y} \mid A = a', \boldsymbol{Z} = \boldsymbol{z}]$. Equivalently, a model is counterfactually invariant if it holds
\begin{equation*}
\expect{\left (\mathbb{E}[\hat{Y}\mid A, \boldsymbol{Z}] - \mathbb{E}[\hat{Y}\mid \boldsymbol{Z}] \right )^2} = 0,
\end{equation*}
where the outer expected value is taken on $A$ and $\boldsymbol{Z}$. By the tower property of the expectation~\citep{Williams-1991}, we have that
\begin{align*}
    \expect{\left (\mathbb{E}[\hat{Y}\mid A, \boldsymbol{Z}] - \mathbb{E}[\hat{Y}\mid \boldsymbol{Z}] \right )^2} 
    & = \expect{\expect{\left (\mathbb{E}[\hat{Y}\mid A, \boldsymbol{Z}] - \mathbb{E}[\hat{Y}\mid \boldsymbol{Z}] \right )^2\mid \boldsymbol{Z}}} \\
    & = \expect{\expect{(\mathbb{E}[\hat{Y}\mid A, \boldsymbol{Z}]^2 - \mathbb{E}[\hat{Y}\mid A, \boldsymbol{Z} ] \mathbb{E}[Y\mid \boldsymbol{Z}])\mid \boldsymbol{Z}}} \\
    & = \expect{\mathbb{E}[\hat{Y}\mid A, \boldsymbol{Z}]^2} - \expect{\expect{\mathbb{E}[\hat{Y}\mid A, \boldsymbol{Z}]  \mathbb{E}[\hat{Y}\mid \boldsymbol{Z}]\mid \boldsymbol{Z}}}\\
    & = \expect{\mathbb{E}[\hat{Y}\mid A, \boldsymbol{Z}]^2} - \expect{\mathbb{E}[\hat{Y}\mid \boldsymbol{Z}]^2}\\
    & = \expect{\hat{Y}\mathbb{E}[\hat{Y}\mid A, \boldsymbol{Z}]} - \expect{\hat{Y}\mathbb{E}[\hat{Y}\mid \boldsymbol{Z}]}.
\end{align*}
The claim follows, since for a classifier $f$ with $\{c_1, \dots, c_n\}$ classes and a discrete random variable $A$, we have that $f$ is counterfactually invariant if and only if it holds
\begin{align*}
\mathbb{E} \left [ f(\hat{x}_{a}) \cdot \mathsf{Avg} (\boldsymbol{z}, a) \right ] & = \expect{\hat{Y}\mathbb{E}[\hat{Y}\mid A, \boldsymbol{Z}]} \\
& = \expect{\hat{Y}\mathbb{E}[\hat{Y}\mid \boldsymbol{Z}]} \\
& = \sum_{\hat{a}\in A} \mathbb{E} \left [ f(\hat{x}_{\hat{a}}) \cdot \mathsf{Avg} (\boldsymbol{z}, \hat{a}) \right ],
\end{align*}
as claimed.
\end{proof}
\section{Datasets description}
\subsection{Synthetic datasets}
\label{app:synthetic_data_description}
We consider six synthetic datasets, namely, $\mathsf{linear}$, $\mathsf{quadratic}$, $\mathsf{exponential}$, $\mathsf{interactive}$, $\mathsf{log}$-$\mathsf{exponent}$, and $\mathsf{sin}$. Each synthetic datasets consist of features $\boldsymbol Z = \{ Z_0, \dots, Z_n\}$ together with a protected feature $A$, as well as observations $X$ and corresponding labels $Y$. The DGP is described by the following equations:
\begin{equation*}
   \label{eq:syntheticdgp}
\left \{
\begin{array}{ll}
   A & \sim \textsc{Bernoulli}(0.3)\\
    Z_{i} &= M_{i}Z_{i+1}A + N_{i}Z_{i+1}(1-A) + \xi_{i}, \ t=0,1,..,n-1 \\
    X &= M_{X} Z_{0} + r + \epsilon_{X} \\
    Y & \sim \textsc{Bernoulli}(\textsc{Sigmoid}(f_{Y}(Z_{n})))
\end{array}
\right .
\end{equation*}
In this data generating process, parameters $M_{i},N_{i},M_{X} \in \mathbb{R}^{k\times k}$ are transformation matrices; $r \in \mathbb{R}^{k}$ is a parameter representing the bias; $\xi_{i},\epsilon_{X} \sim \mathcal{N}(0, \sigma^{2})$ are random errors. $f_{Y}$ is a transformation layer that adds nonlinearity before the sigmoid activation. For each transformation, we use a different random realization of the parameters, thereby generating different synthetic datasets. We apply one linear transformation and five nonlinear transformations for $f_{Y}$:
\begin{itemize}
    \item $\mathsf{linear}$: $ f_{Y}(Z_{n}) = \omega_{Y}\cdot Z_{n} + b_{Y}$, $\omega_{Y} \sim \mathcal{N}(\boldsymbol{0}, I)$, $b_{Y} \sim \mathcal{N}(0, 1)$. 
    \item $\mathsf{quadratic}$: $ f_{Y}(Z_{n}) = \omega_{Y}\cdot (Z_{n})^2 + b_{Y}$, $\omega_{Y} \sim \mathcal{N}(\boldsymbol{0}, 4I)$, $b_{Y} \sim \mathcal{N}(20, 1)$. 
    \item $\mathsf{exponential}$: $f_{Y}(Z_{n}) = \omega_{Y}\cdot e^{Z_{n}} + b_{Y}$, $\omega_{Y} \sim \mathcal{N}(\boldsymbol{0}, I)$, $b_{Y} \sim \mathcal{N}(10, 1)$. 
    \item $\mathsf{interactive}$: $f_{Y}(Z_{n}) = \sum_{i \neq j} \omega_{Y}^{i,j}\cdot Z_{n}^{i}\cdot Z_{n}^{j} + b_{Y}$, $\omega_{Y}^{i,j} \sim \mathcal{N}(0, 1)$, $b_{Y} \sim \mathcal{N}(0, 1)$.    
    \item $\mathsf{log}$-$\mathsf{exponent}$: $f_{Y}(Z_{n}) = \log(\exp(\omega_{Y} \cdot Z_{n}) + b_{Y})$, $\omega_{Y} \sim \mathcal{N}(0, I)$, $b_{Y} \sim \mathcal{N}(5, 1)$. 
    \item $\mathsf{sin}$: $f_{Y}(Z_{n}) = \omega_{Y}\cdot \sin(Z_{n}) + b_{Y}$, $\omega_{Y} \sim \mathcal{N}(\boldsymbol{0}, I)$, $b_{Y} \sim \mathcal{N}(2, 1)$. 
\end{itemize}
Here, $\omega_{Y}$ and $b_{Y}$ are fixed model parameters during the generation process. During the simulation, we take the following values on the hyperparameters $T=3, k=32, \sigma=0.01$. The SCM's parameter $M_{t},N_{t}$ are random matrix whose elements are sampled from uniform distribution $[-10, 10]$. The initial state $Z_{n} \sim \mathcal{N}(0, I_{k})$. {For reproducibility, we publish the synthetic data on Zenodo (\href{https://doi.org/10.5281/zenodo.18803863}{https://doi.org/10.5281/zenodo.18803863}).}
\subsection{The \textsc{CheXpert} dataset}
{\textsc{CheXpert} is a large-scale dataset comprising $224,316$ chest radiographs from $65,240$ patients \citep{DBLP:conf/aaai/IrvinRKYCCMHBSS19}. Each image is annotated for the presence of common chest radiographic observations. Following previous work \citep{DBLP:conf/iccv/Yuan0SY21}, we focus on five of these observations in our experiments: Cardiomelagy, Edema, Consolidation, Atelectasis, and Pleural Effusion. The \textsc{CheXpert} dataset also provides demographic labels, including self-reported race and sex. Overall, $55.5\%$ of patients are labeled as \say{male} and $44.5\%$ as \say{female}. Regarding race, $4.8\%$ of patients are labeled as \say{black}, $10.8\%$ as \say{asian}, $56.2\%$ as \say{white}, and $28.2\%$ as \say{unknown}. For our experiments, we filter the \textsc{CheXpert} dataset to include only frontal chest images and restrict the racial categories to \say{black}, \say{white}, and \say{asian}.}

\subsection{The MIMIC-CXR dataset}
{\textsc{MIMIC-CXR} \citep{mimic-cxr, mimic-cxr-jpg} is a large, publicly available dataset comprising chest X-ray images and their corresponding radiological reports. Similar to the \textsc{CheXpert} dataset, image labels are automatically extracted from the reports, and we focus on 5 out of the 14 available observations: Cardiomegaly, Edema, Consolidation, Atelectasis, and Pleural Effusion. The dataset also includes patient demographic information, such as sex and self-reported race, providing a research setting consistent with that of \textsc{CheXpert}. In this dataset, $52.17\%$ of patients are labeled as \say{male} and $47.83\%$ as \say{female}. Regarding race, $3.2\%$ of patients are labeled as \say{asian}, $18.5\%$ as \say{black}, and $67.6\%$ as \say{white}. For our experiments, we restrict the dataset to patients labeled as \say{black}, \say{white}, or \say{asian}, and include only frontal-view images.}

\section{Base classifiers for experimental comparison}
\subsection{Base classifiers for the synthetic datasets}
\label{app:synthetic_classifiers}
In our experiment the examined objects are classifiers that predicts $Y$ given $X$. We create a diverse model pool containing 1000 statistical learning models. To achieve this, we first define several base model architectures, such as Support Vector Classifiers, Logistic Regression, Decision Trees, and ensemble methods. For each of the base architectures, we establish a comprehensive hyperparameter configuration, which is shown in Table~\ref{tab:syn-classifier}. This process results in a large set of unique ML model configurations (over 400 distinct combinations). The final pool of 1000 instantiated models is then created by sampling from these configurations and setting random seeds. The implementation is directly sourced from the scikit-learn package. 

\begin{table}[t]
\centering
\footnotesize
\caption{\textbf{Machine learning models and their hyperparameters used for classification on the synthetic datasets.}}
\label{tab:syn-classifier}
\begin{tabular}{|l|p{7.5cm}|}
\hline
\textbf{Model} & \textbf{Hyperparameter Search Space} \\ \hline

Support Vector Classifier (SVC) & For Linear Kernel (`kernel='linear'`): \newline $C \in \{0.01, 0.1, 1, 10, 100\}$ \newline \textit{(Total Combinations: 5)} \newline \newline
For RBF Kernel (`kernel='rbf'`): \newline $C \in \{0.1, 1, 10, 100\}$ \newline $\gamma \in \{0.001, 0.01, 0.1, 1, \text{'scale'}\}$ \newline \textit{(Total Combinations: 20)} \newline \newline
For Polynomial Kernel (`kernel='poly'`): \newline $C \in \{0.1, 1, 10, 100\}$ \newline $\text{degree} \in \{2, 3, 4\}$ \newline $\gamma \in \{\text{'scale', 'auto'}\}$ \newline \textit{(Total Combinations: 24)} \\ \hline

Logistic Regression & $C \in \{0.01, 0.1, 1, 10, 100\}$ \newline $\text{penalty} \in \{\text{'l1', 'l2'}\}$ \newline $\text{solver} \in \{\text{'liblinear', 'saga'}\}$ \newline \textit{(Total Combinations: 20)} \\ \hline

Decision Tree Classifier & $\text{criterion} \in \{\text{'gini', 'entropy'}\}$ \newline $\text{max\_depth} \in \{5, 10, 20, 30, \text{None}\}$ \newline $\text{min\_samples\_split} \in \{2, 5, 10\}$ \newline $\text{min\_samples\_leaf} \in \{1, 2, 4\}$ \newline \textit{(Total Combinations: 90)} \\ \hline

Random Forest Classifier & $\text{n\_estimators} \in \{50, 100, 200\}$ \newline $\text{max\_features} \in \{\text{'sqrt', 'log2'}\}$ \newline $\text{max\_depth} \in \{10, 20, \text{None}\}$ \newline $\text{min\_samples\_split} \in \{2, 5\}$ \newline $\text{min\_samples\_leaf} \in \{1, 2\}$ \newline \textit{(Total Combinations: 72)} \\ \hline

Gradient Boosting Classifier & $\text{n\_estimators} \in \{50, 100, 200\}$ \newline $\text{learning\_rate} \in \{0.01, 0.1, 0.2\}$ \newline $\text{max\_depth} \in \{3, 5, 8\}$ \newline $\text{subsample} \in \{0.7, 0.8, 1.0\}$ \newline \textit{(Total Combinations: 81)} \\ \hline

MLP Classifier & $\text{hidden\_layer\_sizes} \in \{(16, 4), (16, 8, 4),$\newline $(32, 16), (50,), (100, 50)\}$ \newline $\text{activation} \in \{\text{'relu', 'tanh'}\}$ \newline $\text{solver} \in \{\text{'adam', 'sgd'}\}$ \newline $\text{alpha} \in \{0.0001, 0.001, 0.01\}$ \newline \textit{(Total Combinations: 90)} \\ \hline

\end{tabular}
\end{table}

\subsection{Base classifiers for \textsc{CheXpert} and MIMIC-CXR}
\label{app:CheXpert_classifiers}
\paragraph{Model architectures.} Same setting is applied to \textsc{CheXpert} and MIMIC-CXR dataset. The diagnosis models, namely the disease classifiers, are trained to predict five diseases simultaneously. The five diseases are cardiomegaly, edema, consolidation and pleural effusion. All the models have the same high level structure: the image is processed through a backbone to produce high-level feature maps. The feature map then goes through a global average pooling operation, reducing the spatial dimension to $(1, 1)$. Then for each of the five labels, dropout and batch normalization are applied before passing through a label-spECAfic $1\times 1$ convolutional classifier to obtain the logits. We consider three types of CNN models, namely Densenet\citep{densenet}, VGG \citep{vgg} and inception-V3\citep{szegedy2015rethinkinginceptionarchitecturecomputer}. For each type of backbone we select multiple variations. For Densenet, we select Densenet 121, Densenet 169 and Densenet 201. For VGG, we select VGG13, VGG13+BN, VGG16, VGG16+BN, VGG19, VGG19+BN. For VGG family's models, '+BN' means that an additional batch normalization is applied in each convolution layer.
\paragraph{Training details.} The training, validation and test set are splitted in the same way as we did in training the DCLDM for CheXpert and MIMIC-CXR respectively. Images are resized to $256\times256$ with the pixel value scaled to $[-1, 1]$. For each model, we download the pretrained backbone from torchvision and then train the full model with a batchsize of 64 and a learning rate of $1e^{-4}$ for $20$ epochs. The loss is computed as the sum of binary cross entropy loss of five labels. We use Adam optimizer. 

\noindent{\textbf{Performance across demographic groups.} We report the performance of the diagnostic models by describing the ROC-AUC scores. Results are shown in Table \ref{tab:rocauc_results} (CheXpert) and Table \ref{tab:rocauc_results_MIMIC} (MIMIC-CXR).}
\begin{table}[t]
\centering
\footnotesize
\caption{{\textbf{ROC-AUC Scores of $10$ classification models for five diseases across demographic groups on the CheXpert dataset}}}
\label{tab:rocauc_results}
\resizebox{\textwidth}{!}{%
\begin{tabular}{|c|c|c c c c c|}
\hline
\textbf{Model} & \textbf{Race} & \textbf{Cardiomegaly} & \textbf{Edema} & \textbf{Consolidation} & \textbf{Atelectasis} & \textbf{Pleural Effusion} \\
\hline
\multirow{3}{*}{densenet121} & Black  & 0.820 & 0.710 & 0.961 & 0.770 & 0.830 \\
                             & White  & 0.880 & 0.780 & 0.943 & 0.720 & 0.870 \\
                             & Asian  & 0.810 & 0.730 & 0.925 & 0.770 & 0.850 \\
\hline
\multirow{3}{*}{densenet169} & Black  & 0.870 & 0.870 & 0.950 & 0.830 & 0.850 \\
                             & White  & 0.953 & 0.872 & 0.941 & 0.810 & 0.830 \\
                             & Asian  & 0.892 & 0.880 & 0.931 & 0.840 & 0.850 \\
\hline
\multirow{3}{*}{densenet201} & Black  & 0.840 & 0.850 & 0.944 & 0.800 & 0.900 \\
                             & White  & 0.940 & 0.871 & 0.950 & 0.800 & 0.870 \\
                             & Asian  & 0.881 & 0.884 & 0.926 & 0.820 & 0.860 \\
\hline
\multirow{3}{*}{VGG13}       & Black  & 0.860 & 0.850 & 0.961 & 0.830 & 0.890 \\
                             & White  & 0.930 & 0.900 & 0.930 & 0.750 & 0.870 \\
                             & Asian  & 0.880 & 0.949 & 0.918 & 0.790 & 0.830 \\
\hline
\multirow{3}{*}{VGG13+BN}    & Black  & 0.850 & 0.830 & 0.963 & 0.810 & 0.900 \\
                             & White  & 0.890 & 0.820 & 0.931 & 0.810 & 0.910 \\
                             & Asian  & 0.860 & 0.830 & 0.926 & 0.850 & 0.930 \\
\hline
\multirow{3}{*}{VGG16}       & Black  & 0.830 & 0.860 & 0.964 & 0.790 & 0.900 \\
                             & White  & 0.890 & 0.850 & 0.960 & 0.800 & 0.860 \\
                             & Asian  & 0.860 & 0.870 & 0.926 & 0.760 & 0.850 \\
\hline
\multirow{3}{*}{VGG16+BN}    & Black  & 0.820 & 0.870 & 0.950 & 0.840 & 0.880 \\
                             & White  & 0.900 & 0.920 & 0.921 & 0.820 & 0.860 \\
                             & Asian  & 0.870 & 0.910 & 0.911 & 0.840 & 0.870 \\
\hline
\multirow{3}{*}{VGG19}       & Black  & 0.870 & 0.850 & 0.971 & 0.840 & 0.830 \\
                             & White  & 0.940 & 0.910 & 0.930 & 0.810 & 0.810 \\
                             & Asian  & 0.890 & 0.920 & 0.930 & 0.850 & 0.830 \\
\hline
\multirow{3}{*}{VGG19+BN}    & Black  & 0.840 & 0.870 & 0.951 & 0.810 & 0.900 \\
                             & White  & 0.900 & 0.920 & 0.940 & 0.810 & 0.870 \\
                             & Asian  & 0.870 & 0.920 & 0.928 & 0.800 & 0.830 \\
\hline
\multirow{3}{*}{Inception V3} & Black & 0.830 & 0.850 & 0.931 & 0.840 & 0.910 \\
                              & White & 0.890 & 0.860 & 0.940 & 0.790 & 0.870 \\
                              & Asian & 0.850 & 0.891 & 0.926 & 0.820 & 0.840 \\
\hline
\end{tabular}
}
\end{table}

\begin{table}[htbp]
\centering
\footnotesize
\caption{{\textbf{ROC-AUC Scores of $10$ classification models for five diseases across demographic groups on MIMIC-CXR}}}
\label{tab:rocauc_results_MIMIC}
\resizebox{\textwidth}{!}{%
\begin{tabular}{|c|c|c c c c c|}
\hline
\textbf{Model} & \textbf{Race} & \textbf{Cardiomegaly} & \textbf{Edema} & \textbf{Consolidation} & \textbf{Atelectasis} & \textbf{Pleural Effusion} \\
\hline
\multirow{3}{*}{densenet121} & Black  & 0.792 & 0.747 & 0.962 & 0.825 & 0.853 \\
                             & White  & 0.854 & 0.819 & 0.944 & 0.775 & 0.887 \\
                             & Asian  & 0.780 & 0.754 & 0.928 & 0.830 & 0.864 \\
\hline
\multirow{3}{*}{densenet169} & Black  & 0.844 & 0.915 & 0.949 & 0.875 & 0.873 \\
                             & White  & 0.926 & 0.915 & 0.939 & 0.862 & 0.857 \\
                             & Asian  & 0.860 & 0.916 & 0.927 & 0.900 & 0.875 \\
\hline
\multirow{3}{*}{densenet201} & Black  & 0.820 & 0.896 & 0.944 & 0.852 & 0.918 \\
                             & White  & 0.912 & 0.913 & 0.951 & 0.861 & 0.898 \\
                             & Asian  & 0.850 & 0.930 & 0.926 & 0.875 & 0.875 \\
\hline
\multirow{3}{*}{VGG13}       & Black  & 0.832 & 0.892 & 0.960 & 0.872 & 0.905 \\
                             & White  & 0.898 & 0.946 & 0.931 & 0.810 & 0.898 \\
                             & Asian  & 0.852 & 0.990 & 0.917 & 0.854 & 0.863 \\
\hline
\multirow{3}{*}{VGG13+BN}    & Black  & 0.821 & 0.872 & 0.964 & 0.865 & 0.916 \\
                             & White  & 0.857 & 0.867 & 0.930 & 0.869 & 0.943 \\
                             & Asian  & 0.831 & 0.871 & 0.926 & 0.898 & 0.968 \\
\hline
\multirow{3}{*}{VGG16}       & Black  & 0.806 & 0.896 & 0.965 & 0.851 & 0.920 \\
                             & White  & 0.853 & 0.888 & 0.960 & 0.865 & 0.897 \\
                             & Asian  & 0.826 & 0.905 & 0.928 & 0.822 & 0.868 \\
\hline
\multirow{3}{*}{VGG16+BN}    & Black  & 0.792 & 0.914 & 0.949 & 0.887 & 0.897 \\
                             & White  & 0.872 & 0.966 & 0.920 & 0.887 & 0.897 \\
                             & Asian  & 0.847 & 0.952 & 0.913 & 0.877 & 0.892 \\
\hline
\multirow{3}{*}{VGG19}       & Black  & 0.841 & 0.898 & 0.970 & 0.887 & 0.858 \\
                             & White  & 0.912 & 0.957 & 0.932 & 0.862 & 0.856 \\
                             & Asian  & 0.861 & 0.967 & 0.930 & 0.896 & 0.854 \\
\hline
\multirow{3}{*}{VGG19+BN}    & Black  & 0.818 & 0.911 & 0.949 & 0.857 & 0.916 \\
                             & White  & 0.871 & 0.960 & 0.939 & 0.864 & 0.902 \\
                             & Asian  & 0.841 & 0.961 & 0.930 & 0.877 & 0.857 \\
\hline
\multirow{3}{*}{Inception V3} & Black & 0.806 & 0.894 & 0.930 & 0.908 & 0.931 \\
                              & White & 0.861 & 0.902 & 0.941 & 0.854 & 0.911 \\
                              & Asian & 0.817 & 0.926 & 0.927 & 0.870 & 0.867 \\
\hline
\end{tabular}
}
\end{table}

\begin{table}[t]
\caption{\textbf{Network architecture for $m_{\delta}$.}}
\label{tab:m_architecture}
\centering
\begin{tabular}{|l|l|l|}
\hline
\textbf{Layer/Block}        & \textbf{Module}                                                                                   & \textbf{Output Shape}   \\ \hline
\textbf{Input}              & Input Tensor                                                                                      & (32, 32, 8)            \\ \hline
\textbf{Conv Block 1}       & \begin{tabular}[c]{@{}l@{}}Conv2D(16, 3x3) + ReLU \\ BatchNorm \\ Conv2D(16, 3x3) + ReLU \\ BatchNorm \\ MaxPooling(2x2)\end{tabular} & (16, 16, 16)           \\ \hline
\textbf{Conv Block 2}       & \begin{tabular}[c]{@{}l@{}}Conv2D(32, 3x3) + ReLU \\ BatchNorm \\ Conv2D(32, 3x3) + ReLU \\ BatchNorm \\ MaxPooling(2x2)\end{tabular} & (8, 8, 32)             \\ \hline
\textbf{Conv Block 3}       & \begin{tabular}[c]{@{}l@{}}Conv2D(64, 3x3) + ReLU \\ BatchNorm \\ Conv2D(64, 3x3) + ReLU \\ BatchNorm \\ MaxPooling(2x2)\end{tabular} & (4, 4, 64)             \\ \hline
\textbf{Upsample Block 1}   & \begin{tabular}[c]{@{}l@{}}Upsample(2x2) \\ Conv2D(32, 3x3) + ReLU \\ BatchNorm \\ Conv2D(32, 3x3) + ReLU \\ BatchNorm\end{tabular}   & (8, 8, 32)             \\ \hline
\textbf{Upsample Block 2}   & \begin{tabular}[c]{@{}l@{}}Upsample(2x2) \\ Conv2D(16, 3x3) + ReLU \\ BatchNorm \\ Conv2D(16, 3x3) + ReLU \\ BatchNorm\end{tabular}   & (16, 16, 16)           \\ \hline
\textbf{Upsample Block 3}   & \begin{tabular}[c]{@{}l@{}}Upsample(2x2) \\ Conv2D(8, 3x3) + ReLU \\ BatchNorm \\ Conv2D(8, 3x3) + ReLU \\ BatchNorm\end{tabular}     & (32, 32, 8)            \\ \hline
\textbf{Final Output}       & Output Tensor                                                                                     & (32, 32, 8)            \\ \hline
\end{tabular}
\end{table}
\section{Implementation details}

\subsection{Conditional Latent Diffusion Models} \label{app:cldm}

\noindent\textbf{Model architecture.} The DCLDM consists of two components, the outside structure VAE and the inner structure diffusion model. The encoder is constructed as a series of convolutional layers and downsampling modules, which progressively reduce the spatial dimensions while capturing essential features from the input data. The decoder, on the other hand, consists of convolutional layers and upsampling modules, designed to reconstruct the original data from the latent representation. To enhance the model's performance, attention mechanisms and skip connections are employed. The DDPM uses U-net as the noise estimator $\epsilon_{\theta}(\boldsymbol{z}_{t}, t, a)$ to denoise the latent space. The U-net is further augmented by a cross-attention mechanism to enable layer propagation with conditioned label. Technically, we adopt the classic archilatent diffusion tecture used for medical images\citep{R-beatsGAN}.
\paragraph{Training details.} The outside VAE structure and the diffusion model inner structure are trained in two successive phases. In the first phase, the VAE is trained to encode a $256\times 256 \times 1$ image into a $32\times 32\times 8$ latent representation and then decode it back into the original image space. Here the number of channels of the latent code adopts the strategy in \citet{M_ller_Franzes_2023} which shows good performance. A multi-resolution loss function, combined with KL-regularization, is used as the training objective. In the second phase, the VAE is frozen, and the input image is encoded into the latent space by the encoder. Random noise is then sampled and used to diffuse the noised latent vector over $1000$ steps. The U-Net \citep{u-net} is subsequently trained to predict the noise from the diffused sample, following the standard training procedure of the diffusion model. During sampling, the diffusion model strategy \citep{DDIM} is applied to accelerate the iterative process.
\paragraph{Variance control.}
For diffusion models, the hyperparameter $\beta_{t}$, $\alpha_{t}=1 - \beta_{t}$ and the number of steps $T$ controls the noise's variance during the forward process, where data is gradually corrupted by adding Gaussian noise. $\beta_{t}$ governs how much noise is added. In settings where the goal is to generate diverse images, the number of steps and $\beta_{t}$ are usually set to $1000$ and a linear interpolation between $0.001$ to $0.01$. In the counterfactual image generation, however, large $\beta$ and $T$ could lead to excessive variance in the generated images. To overcome this problem, in our analysis we use a variance reduction technique. Empirical studies \citet{IDDPM} show that steps between $50$ to $500$ can still produce high-quality images with reduced variance. Therefore in our task, we actively reduce $\beta$ to 20\% and $T$ to $250$ to fit our counterfactual demand. Experiments show that diffusion models achieves lower sample variance without significantly compromising the quality of the generated images.
\begin{table}[t]
\centering
\caption{\textbf{Network architecture for $T_\theta$.}}
\label{tab:mine-architecture}
\begin{tabular}{|l|l|l|}
\hline
\textbf{Layer/Block}        & \textbf{Module}                                                                                       & \textbf{Output Shape}   \\ \hline
\textbf{Input}              & Input Tensor = Concat($\boldsymbol{z}_{0},Emb(a)$)                                                                                          & (32, 32, 10)           \\ \hline
\textbf{Conv1}              & \begin{tabular}[c]{@{}l@{}}Conv2D(10, 64, 3x3, stride=1, padding=1) \\ BatchNorm \\ ReLU\end{tabular}  & (32, 32, 64)           \\ \hline
\textbf{Layer 1}            & \begin{tabular}[c]{@{}l@{}}2 x Residual Block \\ (Conv2D(64, 64, 3x3, stride=1) \\ BatchNorm \\ ReLU)\end{tabular} & (32, 32, 64)           \\ \hline
\textbf{Layer 2}            & \begin{tabular}[c]{@{}l@{}}2 x Residual Block \\ (Conv2D(64, 128, 3x3, stride=2) \\ BatchNorm \\ ReLU)\end{tabular} & (16, 16, 128)          \\ \hline
\textbf{Layer 3}            & \begin{tabular}[c]{@{}l@{}}2 x Residual Block \\ (Conv2D(128, 256, 3x3, stride=2) \\ BatchNorm \\ ReLU)\end{tabular} & (8, 8, 256)            \\ \hline
\textbf{Layer 4}            & \begin{tabular}[c]{@{}l@{}}2 x Residual Block \\ (Conv2D(256, 512, 3x3, stride=2) \\ BatchNorm \\ ReLU)\end{tabular} & (4, 4, 512)            \\ \hline
\textbf{Global Avg Pool}    & AdaptiveAvgPool2d(1x1)                                                                                 & (1, 1, 512)            \\ \hline
\textbf{Fully Connected}    & Linear(512, 1)                                                                                         & (1, 1)                \\ \hline
\textbf{Final Output}       & Real Number Output                                                                                     & (1)                   \\ \hline
\end{tabular}
\end{table}
\subsection{Learning the disentangling transformation} \label{app:transformation}
To improve performance, we remove the influence of the original sensitive attribute (e.g., self-reported race) from the patient's latent representation $\boldsymbol{z}_0$. We achieve this by learning a non-linear transformation $\phi_{\lambda,\delta}$ that makes the resulting representation $\tilde{\boldsymbol{z}}_0 = \phi_{\lambda,\delta}(\boldsymbol{z}_0)$ disentangled from the observed attribute $A$.

The optimal transformation is one that minimizes the Mutual Information (MI) between the transformed representation and the attribute, $I(\phi_{\lambda, \delta}(\boldsymbol{z}_{0}), A)$. Since calculating MI directly is intractable for high-dimensional data, we instead optimize a tractable lower-bound using the Mutual Information Neural Estimation (MINE) framework \citep{belghazi2021minemutualinformationneural}. This turns the problem into a min-max optimization game between the transformation function and a neural network that estimates the MI:
\begin{equation}
\min_{\lambda,\delta} \max_{\theta} \mathbb{E} [\FGen_\theta(\phi_{\lambda, \delta}(\boldsymbol{z}_{0}), a)] - \log\left (\mathbb{E}_{\boldsymbol{Z}} \left [\mathbb{E}_{A}[e^{\FGen_\theta(\phi_{\lambda, \delta}(\boldsymbol{z}_{0}), a)}] \right ]\right ) \label{eq:neural_mutual_info_app}
\end{equation}
where $\FGen_\theta$ is the MINE estimator network. We solve this objective with an alternating optimization scheme. Below we describe the implementation of the two core components: the MINE estimator $\FGen_\theta$ and the transformation $\phi_{\lambda, \delta}$.
\paragraph{Implementation of $T_\theta$.}
The MINE framework relies on a neural network, which we denote $T_\theta$, to provide a tight lower-bound on the mutual information. This network takes a (representation, attribute) pair as input and outputs a single scalar value.

\begin{itemize}
    \item \textbf{Architecture:} The network $T_\theta$ uses a ResNet-18-style architecture. The latent representation $\boldsymbol{z}_0$ (shape: 32x32x8) is first concatenated with a one-hot encoding of the attribute $a$ that has been tiled to match the spatial dimensions (shape: 32x32x2), resulting in an input tensor of shape 32x32x10. This tensor is processed through a series of residual blocks, followed by a global average pooling layer and a final fully connected layer to produce the scalar output. The full architecture is detailed in Table \ref{tab:mine-architecture}.
    \item \textbf{Training:} In the min-max optimization, the parameters $\theta$ of this network are updated to maximize the objective in Eq. \eqref{eq:neural_mutual_info_app}. We use the Adam optimizer with a learning rate of $1e-4$. For every one update of the transformation network, we update the MINE estimator five times to ensure the MI estimate remains tight.
\end{itemize}
\paragraph{Implementation of the transformation function $\phi_{\lambda,\delta}$.} The transformation $\phi$ is designed to minimally perturb the original latent representation $\boldsymbol{z}_0$ to remove the information about attribute $A$. The transformation is defined as:
$$\phi_{\lambda, \delta}(\boldsymbol{z}_{0}) = \boldsymbol{z}_{0} + \lambda \frac{ m_\delta(\boldsymbol{z}_{0})}{\| m_\delta(\boldsymbol{z}_{0})\|}$$
Here, $m_\delta$ is a neural network that predicts the optimal direction of perturbation in the latent space, and $\lambda$ is a scalar hyperparameter that controls the magnitude of this perturbation.

\begin{itemize}
    \item \textbf{Architecture:} The perturbation network $m_\delta$ must take a latent vector $\boldsymbol{z}_0$ and output a perturbation of the same dimension. We therefore employ a U-Net-like CNN architecture, which is well-suited for such image-to-image translation tasks. It consists of an encoder path that downsamples the representation and a decoder path that upsamples it back to the original dimension, using skip connections to preserve fine-grained information. The full architecture is detailed in Table \ref{tab:m_architecture}.
    \item \textbf{Training:} The parameters $\delta$ of the perturbation network are updated to \textit{minimize} the MI estimated by $T_\theta$. We again use the Adam optimizer with a learning rate of $1e-4$. The perturbation magnitude $\lambda$ was set to $1e-3$ based on a random search.
\end{itemize}

\subsection{{Classifier for the self-reported race}}
\label{app:race_classifier}
{The classifiers are instantiated with ResNet-101. Here we report the network architecture for better reproducibility. The network takes as input a grayscale image of size $256\times256\times1$. We adopt the implementation from torchvision package (torchvision.models.resnet101) with the first convolution layer modified to accept 1-channel inputs. After the final global average pooling layer, the 2048-dimensional vector is mapped by a fully connected layer with softmax activation. The optimization uses Adam optimizer with 1e-4 learning rate. The total training epoch number is $30$ and the batch size is  15.}
\end{document}